\documentclass{article}

\usepackage{microtype}
\usepackage{graphicx}
\usepackage{subcaption}
\usepackage{booktabs} % for professional tables

\usepackage{hyperref}

\usepackage[accepted]{icml2024}

\usepackage{url}
\usepackage{amsmath,amssymb,amsthm,bm}
\usepackage{amsthm} 

\newtheorem*{lemma*}{Lemma}
\newtheorem*{claim*}{Claim}

\usepackage{mathtools, nccmath}
\usepackage{multicol,multirow,amsmath,bm}
\usepackage{subcaption}
\usepackage[algo2e]{algorithm2e}
\usepackage{soul}

\usepackage[capitalize,noabbrev]{cleveref}

\theoremstyle{plain}
\newtheorem{theorem}{Theorem}

\newtheorem{lemma}[theorem]{Lemma}
\newtheorem{corollary}[theorem]{Corollary}
\theoremstyle{definition}
\newtheorem{definition}{Definition}
\newtheorem{assumption}{Assumption}
\newtheorem{remark}{Remark}

\usepackage[textsize=tiny]{todonotes}

\usepackage{tikz}
\usepackage{anyfontsize}
\usetikzlibrary{arrows,matrix,positioning,decorations.pathreplacing,calc}
\usepackage{multirow}
\usepackage{hhline}
\allowdisplaybreaks

\newcommand{\policy}{$\mathsf{MOAC}$~}
\newcommand{\policyns}{$\mathsf{MOAC}$}

\usepackage{comment}
\usepackage{enumitem}
\usepackage{wrapfig}

\setenumerate{label=\arabic*)}% global settings

\icmltitlerunning{Finite-Time Convergence and Sample Complexity of Actor-Critic Multi-Objective Reinforcement Learning}

\newcommand{\A}{\mathbf{A}}

\renewcommand{\b}{\mathbf{b}}

\newcommand{\g}{\mathbf{g}}

\newcommand{\J}{\mathbf{J}}

\renewcommand{\r}{\mathbf{r}}

\newcommand{\w}{\mathbf{w}}

\newcommand{\x}{\mathbf{x}}

\begin{document}

\twocolumn[
%\icmltitle{On the Convergence and Sample Complexity of an Actor-Critic Method for Multi-Objective Reinforcement Learning}
\icmltitle{Finite-Time Convergence and Sample Complexity of Actor-Critic Multi-Objective Reinforcement Learning}



\icmlsetsymbol{equal}{*}
\begin{icmlauthorlist}
\icmlauthor{Tianchen Zhou}{equal,osu,amazon}\hspace{-1.5pt}
\icmlauthor{FNU Hairi}{equal,uww}\hspace{-1.5pt}
\icmlauthor{Haibo Yang}{rit}\hspace{-1.5pt}
\icmlauthor{Jia Liu}{osu}\hspace{-1.5pt}
\icmlauthor{Tian Tong}{amazon}\hspace{-1.5pt}
\icmlauthor{Fan Yang}{amazon}\hspace{-1.5pt}
\icmlauthor{Michinari Momma}{amazon}\hspace{-1.5pt}
\icmlauthor{Yan Gao}{amazon}
\end{icmlauthorlist}
\icmlaffiliation{amazon}{Amazon, Seattle, WA, USA}
\icmlaffiliation{uww}{Department of Computer Science, University of Wisconsin Whitewater, WI, USA}
\icmlaffiliation{rit}{Department of Computing and Information Sciences, Rochester Institute of Technology, Rochester, NY, USA}
\icmlaffiliation{osu}{Department of Electrical and Computer Engineering, The Ohio State University, Columbus, OH, USA}

\icmlcorrespondingauthor{Tianchen Zhou}{tiancz@amazon.com}

\icmlkeywords{Machine Learning, ICML}

\vskip 0.3in
]



\printAffiliationsAndNotice{\icmlEqualContribution} % otherwise use the standard text.

\begin{abstract}
Reinforcement learning with multiple, potentially conflicting objectives is pervasive in real-world applications, while this problem remains theoretically under-explored.
This paper tackles the multi-objective reinforcement learning (MORL) problem and introduces an innovative actor-critic algorithm named \policy which finds a policy by iteratively making trade-offs among conflicting reward signals.
Notably, we provide the first analysis of finite-time Pareto-stationary convergence and corresponding sample complexity in both discounted and average reward settings.
Our approach has two salient features: (a) \policy mitigates the cumulative estimation bias resulting from finding an optimal common gradient descent direction out of stochastic samples.
This enables provable convergence rate and sample complexity guarantees independent of the number of objectives;
% This further implies that, with proper parameters, \policy scales comparably to single-objective reinforcement learning;
(b) With proper momentum coefficient, \policy initializes the weights of individual policy gradients using samples from the environment, instead of manual initialization. 
This enhances the practicality and robustness of our algorithm.
Finally, experiments conducted on a real-world dataset validate the effectiveness of our proposed method.

\end{abstract}

\section{Introduction}

{\bf 1) Background and Motivation:}
As a foundational machine learning paradigm, reinforcement learning (RL) is concerned with a sequential decision-making process in which an agent interacts with an environment and repeats the tasks of observing the current state, performing a policy-guided action, and receiving rewards and transitioning to the next state.
Upon collecting a trajectory of action-reward sample pairs, the agent updates its policy to maximize its long-term accumulative reward.
To date, although RL has found a large number of applications (e.g., healthcare \citep{petersen2019deep,raghu2017continuous}, financial recommendation \citep{theocharous2015personalized}, ranking system \citep{wen2023rankitect}, resources management \citep{mao2016resource} and robotics \citep{levine2016end,raghu2017deep}), the standard RL formulation only considers a {\em single} reward optimization.
%(e.g., \citet{yang2019provably,XuWanLia_20,afsar2022reinforcement,ma2020off}).
However, as RL applications with increasingly more complex reward structures emerge, it has become apparent that the single-reward structure in the traditional RL framework is not rich enough to capture the needs of these complex RL applications, particularly those with {\em multiple reward objectives.}

For example, in RL-based short video recommender systems~\cite{CaiXueZha_23}, the agent recommends short videos to optimize a multi-dimensional reward rate that captures users' ``WatchTime,'' ``Like,'' ``Dislike,'' ``Comment,'' etc.
%\kevin{Tianchen, state the Kuaishou short video motivating example here.} 
As another example, e-commerce recommender systems rank and display products by taking into account and balancing the preferences of different user groups, some of which prefer faster delivery, while others may prefer lower prices and tolerate slower delivery. 
All of these new multi-objective RL applications necessitate solving {\em multi-objective reinforcement learning} (MORL) problems~\cite{StaKarAra_22,GeZhaYu_22,CheDuXia_21}.
So far, however, research on MORL is still in its infancy and there remains a lack of rigorous theoretical understanding of MORL algorithmic designs in terms of finite-time convergence and sample complexity analysis.
This gap motivates us to take the first attempt at building a theoretical foundation for MORL in this work.

Formally, the learning of policy parameters for an $M$-objective MORL problem can be formulated as follows:
\begin{align}\label{eqn:MORL}
    \max_{\bm{\theta}\in\mathbb{R}^{d_1}}\bm{J}(\bm{\theta}) := \left( J^1(\bm{\theta}), J^2(\bm{\theta}), \ldots, J^M(\bm{\theta}) \right)^{\top},
\end{align}
where $\bm{\theta}$ is the policy parameter of policy $\pi_{\bm{\theta}}$, and $J^i(\bm{\theta})$ is the expected accumulative reward for objective $i\in\{1,\ldots,M\}$ induced by policy $\pi_{\bm{\theta}}$ (we will focus on two objectives of $J^{i}(\bm{\theta})$ commonly used in RL later in Section \ref{sec:model}).
Note, however, that there may not always exist a common policy parameter $\bm{\theta}$ that maximizes all individual objectives simultaneously in \eqref{eqn:MORL} due to the potential conflicts among the objectives.
Therefore, a more appropriate and relevant metric in MORL is to find a Pareto-optimal solution for all objectives, where no objective can be unilaterally further improved without sacrificing another objective.
%However, similar to single-objective non-convex optimization problem, finding global optimum (or equivalently Pareto optimal in multi-objective setting) is NP-hard in general \citep{DanDvuGas_22,YanLiuLiu_23}. 
Further, due to non-convexity typically manifested in MORL problems in practice, finding Pareto-optimal solution is NP-hard in general~\citep{DanDvuGas_22,YanLiuLiu_23}.
To address this challenge, the notion called {\em Pareto-stationary solution} (a necessary condition for being Pareto-optimal) is commonly adopted in solving non-convex multi-objective optimization problems \citep{Des_12,SenKol_18,YanLiuLiu_23}. 
As a first step towards understanding and characterizing MORL finite-time convergence and sample complexity, we focus on the Pareto-stationary convergence of MORL in this paper.

{\bf 2) Technical Challenges:}
Just as the close relationship between actor-critic policy-gradient approaches \citep{grondman2012survey,kumar2019sample,XuWanLia_20} for RL and the gradient-based methods for general optimization problems, a natural idea to solve Problem~\eqref{eqn:MORL} is to develop an actor-critic policy-gradient MORL method by drawing inspirations from gradient-based multi-objective optimization (MOO) methods.
In the MOO literature, the multi-gradient descent algorithm (MGDA) is a popular approach for finding a Pareto-stationary solution \citep{Des_12}.
MGDA can be viewed as an extension of the classical gradient descent method to MOO, which aims to identify a common descent direction for all objectives in each iteration.
%as a minimum $\ell_2$ norm element in the convex hull of the gradients from all individual objectives.
Also, the convergence of MGDA and its variants have recently been established under different MOO settings, including convex and non-convex objective functions \citep{liu2021stochastic,FerSheLiu_22} and decentralized data \citep{YanLiuLiu_23}, etc.
However, developing an MGDA-type actor-critic policy-gradient method for MORL with provable Pareto-stationary convergence is highly non-trivial due to the following technical challenges:
\vspace{-.1in}
\begin{list}{\labelitemi}{\leftmargin=1.5em \itemindent=-0.0em \itemsep=-.2em}
\item[a)] In actor-critic RL, the actor and critic components approximate the policy and value functions bootstrapped by the Bellman optimality principle.
This implies an intricate dependence between the actor and critic components. 
Moreover, such an actor-critic dependence is further exacerbated by the complex coupling between multiple objectives, rendering conventional MOO convergence analysis inapplicable for actor-critic policy-gradient MORL methods.
As a result, it is unclear whether one can design a multi-objective actor-critic algorithmic framework with provable Pareto-stationary convergence, and if yes, how to characterize its finite-time convergence and sample complexity.
%This allows the incorporation of MGDA in computing a common gradient descent direction of policy parameter.
%However, it is unclear whether such method can achieve Pareto stationary convergence and if so, how to characterize the finite-time convergence and sample complexity.

\item[b)] In actor-critic MORL, both actor and critic have to update their parameters through stochastic approximations due to the finite trajectory-length constraint in practice. 
As a result, actor-critic MORL methods with stochastic MGDA-type updates inevitably introduce cumulative estimation biases in policy parameter updates.
%, since MGDA always finds optimal gradient weights while the gradients are obtained from sampling. 
If not treated carefully, such cumulative estimation biases could significantly diminish MORL performance in reward maximization or even result in a divergence of policy parameter updates.
\end{list}
%\vspace{-1em}

% As a trade-off, one can always transform a MORL problem to a single-objective reinforcement learning problem by pre-specifying the weights \citep{mossalam2016multi, white1980solution}, however, quantifying the weights becomes a new challenge and it remains unclear whether a learnt policy is optimal on balancing conflicting objectives.

% In multi-objective optimization (MOO), a prominent solution is to find a Pareto stationary point for all objectives that cannot be further improved without sacrificing any individual objective.
% In this context, multiple gradient descent algorithm (MGDA) has been proposed.

% Pareto stationary point is a point satisfying certain first-order necessary conditions for being Pareto optimal. To ease the understanding of the solution concepts in multi-objective context, one can compare it with the conventional single-objective optimization. In such parallel comparison, Pareto optimal point corresponds to a global optimal and Pareto stationary point can be similarly understood as stationary point as in conventional single-objective optimization.

\vspace{-.1in}
{\bf 3) Main Contributions:}
In this paper, we overcome the aforementioned challenges and propose a multi-objective actor-critic algorithmic framework with provable finite-time Pareto-stationary convergence and sample complexity guarantees.
%, which is the first of its kind in the literature to our knowledge.
Collectively, our results provide the first building block toward a theoretical foundation for MORL.
Our main contributions are summarized as follows:
% Our key contribution is that we introduce an MGDA-based actor-critic algorithm for MORL with provable Pareto-stationary convergence rate and sample complexity guarantees.
% Our major contributions are summarized as follows:
\vspace{-1em}
\begin{list}{\labelitemi}{\leftmargin=1.5em \itemindent=-0.0em \itemsep=-.2em}
\item We propose a unifying \underline{m}ulti-\underline{o}bjective \underline{a}ctor-\underline{c}ritic algorithmic framework (\policyns) based on MGDA-style policy-gradient update for both (heterogeneous) discounted and average reward settings in MORL. 
Our \policy policy framework offers finite-time convergence and sample complexity of $\mathcal{O}(1/\epsilon^{2})$ for achieving an $\epsilon$-Pareto stationary solution.
To our knowledge, such finite time convergence and sample complexity results are the first of its kind in the MORL literature.
% \item By utilizing momentum approach for weight balance among individual policy gradients and selecting very common step sizes, our algorithm do not require initialization of weights for individual objective gradients. This approach not only ensures finite-time convergence and sample complexity guarantees and also enhances the practicality and robustness of our algorithm and without manually choosing weights among objectives.
\item To mitigate the cumulative estimation bias resulting from stochastic MGDA-type policy parameter update, we propose a momentum mechanism in \policyns. 
The most salient feature of this momentum approach is that the convergence rate and sample complexity of \policy are {\em independent} of the number of objectives. 
This is in a sharp contrast to general MOO, where the convergence results scale either linearly with respect to $M$ \citep{FerSheLiu_22} or even have a high-order dependency $\mathcal{O}(M^{3})$ \citep{ZhoZhaJia_22}. 

\item Based on the proposed momentum mechanism, we show that with a proper schedule of the momentum coefficient, \policy initializes the weights of individual policy gradients out of samples from the environment, instead of manual initialization. This enhances the practicality and robustness of our approach.
%\item We conduct experiment on real-world dataset and show that the off-policy version (for fair comparison) of our method outperforms other relevant state-of-the-art methods in MORL.
\end{list}
\vspace{-1em}
\section{Related Work}
In this section, we provide an overview on two closely related areas, namely MGDA-type MOO methods and MORL, to put our work in comparative perspectives.

\textbf{1) MGDA-type Multi-objective Optimization (MOO):}
Multi-objective optimization \cite{Mie_99} is concerned with optimizing a set of objective functions simultaneously over a set of shared decision variables. 
Compared to conventional single-objective optimization problems, the key difference in MOO is that different objectives could be conflicting.
Thus, the goal of MOO is to determine an equilibrium among the conflicting objectives in the Pareto sense.
Among the approaches for solving MOO problems, the MGDA algorithm \cite{Des_12} has received increasing attention in the learning community in recent years.
However, the convergence analysis in \cite{Des_12} is only asymptotic.
Later in \cite{FliVazVic_19}, it is shown that MGDA achieves the same $\mathcal{O}(1/T)$ finite-time convergence rate as its single-objective counterparts under certain assumptions.
For stochastic MGD (SMGD) methods, the Pareto-stationary convergence analysis is further complicated by the stochastic gradient noise, which often leads to more complex assumptions in their convergence analysis. 
% %and thus it is still unclear whether SMGD is guaranteed to converge.
Notably, an $\mathcal{O}(1/T)$ rate analysis for SMGD was provided in \cite{liu2021stochastic} for strongly convex MOO based on assumptions on a first-moment bound and Lipschtiz continuity of common descent direction.
However, it has been shown in \cite{liu2021stochastic,ZhoZhaJia_22} that the estimation of common descent direction in SMGD could be biased, which may result in divergence.

We note, however, that the convergence analysis in the MOO literature does {\em not} translate into actor-critic MORL methods due to the significant structural differences. 
Specifically, the finite-time convergence rates of MGDA-type MOO methods scale with the number of objectives $M$.
In contrast, we show that the finite-time and sample complexity results of our \policy algorithm are objective-dimension-independent.

\textbf{2) Multi-objective Reinforcement Learning (MORL):}
MORL is a class of sequential decision-making problems with multiple reward signals.
%, e.g. online recommendation services \cite{CheDuXia_21,CaiXueZha_23} and balancing utility and fairness \cite{GeZhaYu_22}. 
Unlike traditional RL problems with scalar-valued rewards (e.g., \citet{SutBar_18,KonTsi_99,XuWanLia_20,GuoHuZha_21}), MORL problems aim at optimizing vector-valued rewards. 
% So far, results on MORL are quite limited in the literature.
% To our knowledge, the first MORL method was reported in \cite{CheDuXia_21}, where an actor-critic MORL algorithm was proposed based on deterministic policy-gradients. 
% Later in \cite{CaiXueZha_23}, a two-stage constrained actor-critic algorithm was proposed. 
Research on MORL can be traced back to 1998 \cite{gabor1998multi,van2014multi,abels2019dynamic,yang2019generalized}.
More recently, \cite{CheDuXia_21} proposes an actor-critic MORL algorithm based on deterministic policy-gradients.
Later in \cite{CaiXueZha_23}, a two-stage constrained actor-critic algorithm was proposed.
We note, however, that the formulation of \cite{CaiXueZha_23} is not in the standard form of MORL, since all but one objective are treated as constraints and the only remaining objective is used as the system objective. 
Moreover, none of these existing works provides finite-time convergence or sample complexity results.

It is worth noting that several RL paradigms also share some similarities with MORL.
For example, in cooperative multi-agent reinforcement learning (MARL) \cite{ZhaYanLiu_18,CheZhoChe_21,HaiLiuLu_22}, each agent has its own (scalar-valued) reward, but the overall objective of cooperative MARL is a {\em fixed} weighted sum of all agents' rewards. 
Similarly, a scalarized version of MORL is considered in \cite{StaKarAra_22}, so that techniques for single-objective RL can be leveraged. 
%utilized for recommender systems. 
Constrained (or safe) RL \cite{WeiLiuYin_22,CaiXueZha_23} is another RL paradigm for balancing multiple RL objectives, where a set of predefined parameters are associated with the constraints to specify the constraint levels. 
%In comparison, the MORL formulation in this work is parameter-free.

\section{A Multi-Objective Actor-Critic Framework}

In this section, we will first introduce the preliminaries of MORL in Section~\ref{sec:model}.
Then, we will present the necessary technical background for defining policy gradients for MORL in Section~\ref{sec:policy_gradient}, which will be useful in describing our proposed \policy algorithmic framework in Section~\ref{sec:policy}.

\subsection{Preliminaries of MORL}
\label{sec:model}

{\bf 1) System Model:}
Consider an RL environment where the instantaneous reward is an $M$-dimensional vector ($M\ge 2$), where each dimension is associated with one objective. 
For convenience, we let $[M]:=\{1,\cdots, M\}$. 
We now formally describe the MORL environment, which is stated as a multi-objective Markov decision process (MOMDP):
\begin{definition}[MOMDP]
A multi-objective Markov decision process is defined by a 4-tuple $(\mathcal{S},\mathcal{A},P,\r)$, where $\mathcal{S}$ is the state space, $\mathcal{A}$ is the action space of the agent, $P:\mathcal{S}\times\mathcal{A}\rightarrow\mathcal{S}$ represents a state transition probabilities, and $\r\in\mathbb{R}^M$ is an $M$-dimensional reward. 
\end{definition}

In this paper, we assume that $\mathcal{S}$ and $\mathcal{A}$ are finite. 
The instantaneous reward $r^{i}(s,a)$ for each objective $i\in [M]$ is deterministic under state $s$ and action $a$.\footnote{For ease of presentation, in this paper, we consider the instantaneous rewards as deterministic given state and action pair. However, the results also hold straightforwardly for stochastic instantaneous rewards.} 
We consider parameterized and stationary policies, i.e., a policy $\pi_{\bm{\theta}}$ is parameterized by $\boldsymbol{\theta}\in\mathbb{R}^{d_1}$, and an agent chooses an action $a$ under state $s$ with probability $\pi_{\bm{\theta}}(a|s) \in [0,1]$.
We consider linear value function approximation in this paper: for each objective $i\in[M]$, the value function $V^i(s;\w^i)$ is approximated as $V^i(s;\w^i) \approx \bm{\phi}(s)^{\top}\w^i, \forall s\in\mathcal{S}$, where  $\w^i\in\mathbb{R}^{d_2}$ are the parameters and $\bm{\phi}(s)\in\mathbb{R}^{d_2}$ is the feature mapping for state $s$.
For simplicity, in this paper, we assume that the same feature mapping is shared among all objectives.
We note that it is straightforward to extend our algorithms and results to cases with objective-dependent feature mappings.

% \subsection{A Composite Objective Function}
% \label{sec: J_def}

{\bf 2) Problem Statement:}
We consider two reward settings: i) average total reward and ii) discounted total reward, both of which have a wide range of applications.
%Our goal is to maximize the rewards in both settings in MORL.
The objectives for these two reward settings are defined as follows:
%In MORL, the objective functions for this two settings are defined in the following.

{\bf 2-a) Average Reward:} In the average total reward setting, each objective $i\in [M]$ is defined as:
\begin{equation*}
    J^{i}(\bm{\theta}):=\lim_{T\to\infty}\frac{1}{T} \mathbb{E}\bigg[ \sum_{t=1}^{T} r^{i}_{t}\bigg].
\end{equation*}
% \begin{align*}
% J^{i}(\bm{\theta})&=\lim_{T\to\infty}\frac{1}{T} \mathbb{E}\bigg[ \sum_{t=0}^{T-1} r^{i}_{t+1}\bigg]\\
% &=\sum_{s\in\mathcal{S}}d_{\bm{\theta}}(s)\sum_{a\in\mathcal{A}}\pi_{\bm{\theta}}(a|s)\cdot R^{i}(s,a),
% \label{eq: obj}
% \end{align*} 
% where $R^{i}(s,a)=\mathbb{E}\left[r^{i}_{t+1}|s_t=s,a_{t}=a\right]$.

{\bf 2-b) Discounted Reward:} In the discounted total reward setting, each objective $i \in [M]$ is defined as:
\begin{equation*}
J^{i}(\bm{\theta}):=\lim_{T\to\infty}\mathbb{E}\bigg[ \sum_{t=1}^{T} (\gamma^{i})^tr^{i}_{t}\bigg],
\end{equation*}
where $\gamma^{i}\!\in\!(0,1)$ is the discount factor for objective $i$.

The goal of MORL is to find a policy $\pi_{\bm{\theta}}$ that jointly maximizes all the objective functions in the long run.
Specifically, given a vector-valued objective function $\J(\bm{\theta}) \in \mathbb{R}^{M}$ for either average total reward or discounted total reward, a policy $\pi_{\bm{\theta}}$ maximizes the following composite objective:
\begin{equation}
\max_{\bm{\theta}\in\mathbb{R}^{d_1}}\J(\bm{\theta}) := \left[ J^1(\bm{\theta}), J^2(\bm{\theta}), \ldots, J^M(\bm{\theta}) \right]^{\top}.
\label{def:obj}
\end{equation}

{\bf 3) Performance Metric:}
To address the potential conflicts among the objectives in $\J(\bm{\theta})$ in MORL, we need the following notions of Pareto-optimality and Pareto-stationarity, which are defined as follows:
\begin{definition}(Pareto Optimality)
    A solution $\x$ dominates solution $\x'$ if and only if $J^i(\x)\geq J^i(\x'), \forall i\in[M]$ and $J^i(\x)> J^i(\x'), \exists i\in[M]$. 
    A solution $\x$ is Pareto-optimal if it is not dominated by any other solution.
\end{definition}
Intuitively, in a Pareto-optimal solution, none of the objectives can be unilaterally further improved without sacrificing another objective.
However, since finding a Pareto-optimal solution for non-convex MORL problems is NP-hard, it is often of practical interest to find an $\epsilon$-Pareto-stationary solution instead, which is defined as follows~\citep{Des_12,SenKol_18,YanLiuLiu_23}:
\begin{definition}($\epsilon$-Pareto Stationarity)
    A solution $\x$ is $\epsilon$-Pareto stationary if there exists $\bm{\lambda}\in\mathbb{R}^M$ such that $\min_{\bm{\lambda}}\|\nabla_{\x}\J(\x)\bm{\lambda}\|_2^2\leq\epsilon$ with $\bm{\lambda}\geq \bm{0}$, $|\bm{\lambda}|_1=1$, and $\epsilon>0$.
\end{definition}
Pareto-stationarity is a necessary condition for a solution to be Pareto optimal.
In particular, for convex MORL, Pareto-stationary solutions are also Pareto-optimal.

%\subsection{Model Assumptions}

\subsection{Policy Gradient for MORL} \label{sec:policy_gradient}

Our proposed \policy algorithmic framework is based on the policy gradient of MORL.
To define policy gradient for MORL, we first state several assumptions for $\pi_{\bm{\theta}}(a|s)$, which guarantees the existence of a stationary distribution of $\{s_t\}_{t\geq 0}$ under any given policy.
\begin{assumption}[MOMDP]
    Given an MOMDP, for any state $s\in\mathcal{S}$, action $a\in\mathcal{A}$, policy parameter $\bm{\theta}\in\mathbb{R}^{d_1}$, we have the following:
    \vspace{-1em}
    \begin{list}{\labelitemi}{\leftmargin=1.8em \itemindent=-1.2em \itemsep=-.2em}
    \item[(a)] The policy function $\pi_{\bm{\theta}}(a|s)\geq 0$ is continuously differentiable with respect to the parameter $\bm{\theta}$;
    \item[(b)] The Markov process $\lbrace s_t\rbrace_{t\geq 0}$ induced by the MOMDP is irreducible and aperiodic, with the transition matrix $P_{\bm{\theta}}=\sum_{a\in\mathcal{A}}\pi_{\bm{\theta}}(a|s)\cdot P(s'|s,a), \forall s,s'\in\mathcal{S}$;
    \item[(c)] The instantaneous reward $r^i_t, i\in[M]$ is non-negative and uniformly bounded by a constant $r_{\max}>0$.
    \end{list}
    \vspace{-1em}
\label{ass:mdp}
\end{assumption}
Assumption \ref{ass:mdp} guarantees that the states have a stationary distribution $d_{\bm{\theta}}(s)$ over $\mathcal{S}$ under any policy $\pi_{\bm{\theta}}$. 
As a result, the Markov chain of state action pair $\{(s_t,a_t)\}_t$ has a stationary distribution $d_{\bm{\theta}}(s)\cdot \pi_{\bm{\theta}}(a|s)$.
Also, (c) is  common in the literature (e.g., \citet{ZhaYanLiu_18,XuWanLia_20,DoaMagRom_19}) and easy to be satisfied in many practical MDP models with finite state and action spaces.

\begin{assumption}[Function Approximation]
    The value function of each objective $i$ is approximated by a linear function: $V^i_{\w}(s) \approx \bm{\phi}(s)^{\top}\w^i, i\in[M]$, where $\w^i\in\mathbb{R}^{d_2}$ with $d_2<|\mathcal{S}|$ is a parameter to be learnt, and $\bm{\phi}(s)\in\mathbb{R}^{d_2}$ is the feature associated with state $s\in\mathcal{S}$, which satisfies:
    \vspace{-1em}
    \begin{list}{\labelitemi}{\leftmargin=1.8em \itemindent=-1.2em \itemsep=-.2em}
    \item[(a)] All features are normalized, i.e., $\|\bm{\phi}(s)\|_2\leq 1, \forall s\in\mathcal{S}$;
    \item[(b)] The feature matrix $\Phi\in\mathbb{R}^{|\mathcal{S}|\times d_2}$ has full column rank;
    \item[(c)] For any $u\in \mathbb{R}^{d_2}$, $\Phi u\neq \mathbf{1}$, where $\mathbf{1}\in\mathbb{R}^{d_2}$;
    \item[(d)] Let $\A_{\pi_{\bm{\theta}}}:=\mathbb{E}_{s\sim d_{\bm{\theta}}(s), s'\sim P(\cdot|s)}[(\bm{\phi}(s')-\bm{\phi}(s))\bm{\phi}^{\top}(s)]$ if in average reward setting. Otherwise, if in discounted reward setting, let $\A_{\pi_{\theta}}:=\mathbb{E}_{s\sim d_{\theta}(s), s'\sim P(\cdot|s)}\left[\left(\gamma\bm{\phi}(s')-\bm{\phi}(s)\right)\bm{\phi}^{T}(s)\right]$. Then, there exists a constant $\lambda_{\A}>0$ such that $\lambda_{\max}(\A_{\pi_{\bm{\theta}}}+\A^{\top}_{\pi_{\bm{\theta}}})\le -\lambda_{\A}$, where $\lambda_{\max}(\A)$ is the largest eigenvalues of the matrix $\A$. % and there a corresponding constant $\lambda_{\A'}>0$ for discounted setting.
    \end{list}
    \vspace{-1em}
\label{ass:feature}
\end{assumption}
Assumption \ref{ass:feature}(c)(d) implies that for any policy $\pi_{\bm{\theta}}$, the inequality $\w^{\top}\A_{\pi_{\bm{\theta}}}\w<0$ holds for any $\w\neq 0$, and $\A_{\pi_{\bm{\theta}}}$ is invertible with $\lambda_{\max}(\A_{\pi_{\bm{\theta}}}+\A^{\top}_{\pi_{\bm{\theta}}})\le 0$.
This ensures that the optimal approximation $\w^*_{\bm{\theta}}$ for any given policy $\pi_{\bm{\theta}}$ is uniformly bounded.
Assumption~\ref{ass:feature} is standard and has been widely use in the literature (e.g., \citet{TsiVan_99,ZhaYanLiu_18,QiuYanYe_21}).

For each objective $i\in [M]$, we define the state-action value function as follows:
(i) for average total reward:
$Q^{i}_{\bm{\theta}}(s,a):=\mathbb{E}\left[\sum_{t=0}^\infty r^{i}(s_t,a_t)-J^i(\bm{\theta})|s_0=s,a_0=a\right]$, 
and (ii) for discounted total reward:
$Q^{i}_{\bm{\theta}}(s,a)=\mathbb{E}\left[\sum_{t=0}^{\infty}(\gamma^i)^t r^{i}(s_t,a_t)|s_0=s,a_0=a,\bm{\theta}\right]$.
It then follows that the value function satisfies:
$V^{i}_{\bm{\theta}}(s)=\sum_{a\in\mathcal{A}}Q^{i}_{\bm{\theta}}(s,a)\cdot\pi_{\bm{\theta}}(a|s).$
We define the advantage function as follows:
\begin{equation*}
\text{Adv}^{i}_{\bm{\theta}}(s,a)=Q^{i}_{\bm{\theta}}(s,a)-V^{i}_{\bm{\theta}}(s), \quad \forall i\in [M]. \label{eq: adv_fun}
\end{equation*}
Let function $\bm{\psi}_{\bm{\theta}}(s,a):=\nabla_{\bm{\theta}}\log \pi_{\bm{\theta}}(a|s)$ be the score function for state-action pair $(s,a)$. 
The gradient of a policy $\pi_{\bm{\theta}}$ for policy gradient is then stated in the following lemma.
\begin{lemma}[Policy Gradient Theorem]{\em
For any $\bm{\theta}$, let $\pi_{\bm{\theta}}:\mathcal{S}\times\mathcal{A}\to[0,1]$ be a policy and let $J^{i}(\bm{\theta})$ be the total reward for the $i$-th objective. 
Then, the policy-gradient of $J^{i}(\bm{\theta})$ with respect to parameter $\bm{\theta}$ can be computed as:
\begin{equation*}
\nabla_{\bm{\theta}}J^{i}(\bm{\theta})=\mathbb{E}_{s\sim d_{\bm{\theta}},a\sim\pi_{\bm{\theta}}}\left[\bm{\psi}_{\bm{\theta}}(s,a)\cdot \textup{Adv}^{i}_{\bm{\theta}}(s,a)\right].
\end{equation*}
\label{the: pol_gra}
}
\end{lemma}
\vspace{-20pt}
We note that Lemma~\ref{the: pol_gra} is a natural extension of the policy gradient in single-objective RL \cite{SutMcASin_99} to any individual objective $i\in [M]$ in MORL.
% However, in order to achieve a update gradient direction that balances all objectives, it requires to consider a common direction that maximizes objective functions $J^{i}(\theta)$ for all $i\in [M]$.
% A common approach among multi-objective gradient descent algorithm (MGDA) \footnote{Although the formulation for our reinforcement learning is maximization problem, we continue to use the more common term MGDA to refer the classic algorithm.} and its variants is to find a common direction via solving the following problem.

% \begin{equation}
%     \min_{\bm{\lambda}\in\mathbb{R}^M}\quad \| \sum_{i=1}^M\lambda^i\cdot \nabla\bm{J}^i(\theta) \|_2^2\quad
%     s.t.\quad \bm{\lambda}\geq \bm{0}, \|\bm{\lambda}\|_1=1, \nonumber
% \end{equation}
% which is a convex quadratic programming with linear constraints. By solving the above problem, one can obtain a common update direction for $\theta$, which is $\bm{d}=\sum_{i\in [M]}\lambda^{i} \nabla\bm{J}^i(\theta)$.
\subsection{The Proposed \policy Algorithmic Framework} \label{sec:policy}

With the preliminaries in Sections~\ref{sec:model} and \ref{sec:policy_gradient}, we are now in a position to present our proposed multi-objective actor-critic (\policyns) algorithmic framework for both average total reward and discounted total reward settings.
\policy iterates over $T$ rounds by alternating between two steps (i) actor (cf. Algorithm~\ref{alg: actor}) and (ii) critic (cf. Algorithm~\ref{alg: critic}) with temporal differential (TD) learning. 
In each iteration, the critic updates $M$ value function estimations through an inner-loop given policy estimation from the previous iteration (cf. Algorithm~\ref{alg: critic}).
Then, with updated value function estimations, the actor updates its policy parameters $\bm{\theta}$ (cf. Algorithm~\ref{alg: actor}).
In both steps, we use constant step-sizes and mini-batch Markovian sampling.

\begin{algorithm}[t]
  \SetKwInOut{Input}{Input}
  \SetKwInOut{Output}{Output}
  \SetKwFor{ParFor}{for}{do in parallel}{end for}
  \Input{$s_0$, $\bm{\theta}_t$, $\Phi$, critic step size $\beta$, critic iteration $N$, critic batch size $D$}
  \BlankLine
  \For{$k=1,\cdots,N$}{
  	$s_{k,1}=s_{k-1,D}$ (when $k=1$, $s_{1,1}=s_0$)
   
	\For{$\tau=1,\cdots,D$}{
        Execute action $a_{k,\tau}\sim\pi_{\bm{\theta}_t}(\cdot|s_{k,\tau})$
        
	Observe state $s_{k,\tau+1}$ and reward vector $\r_{k,\tau+1}$
 
        \ParFor{$i\in [M]$}{
            $\bullet$ {\em Setting I: Average Reward:} 

            Update $\mu_{k,\tau}^i$, $\delta^i_{k,\tau}$ by Eqs.~(\ref{eq:critic_avg1}),(\ref{eq:critic_avg2}), respectively

            \BlankLine
            
            $\bullet$ {\em Setting II: Discounted Reward:} 

            Update $\delta^i_{k,\tau}$ by Eq.~(\ref{eq:critic_disc})
	    }
  }
    \ParFor{$i\in [M]$}{
  $\w^{i}_{k}= \w^{i}_{k-1}+\frac{\beta}{D}\sum_{\tau=1}^{D}\delta^{i}_{k,\tau}\cdot\bm{\phi}(s_{k,\tau})$
  }
  }
  \Output{$\lbrace\w^{i}_{N}\rbrace_{i\in [M]}$, $s_{N,D}$}
  \caption{\policy critic with mini-batch TD-learning.}
  \label{alg: critic}
\end{algorithm}

\textbf{1) The Critic Step:} The critic step is 
%achieved through its own oracle, which is 
illustrated in Algorithm~\ref{alg: critic}.
Given policy $\pi_{\bm{\theta}_t}$, in each iteration $k$ in the critic step, the value function parameters $\w^i_k, i\in[M]$ are locally and concurrently updated through a batch of Markovian samplings.
% In each iteration $k$ in the critic step, the value function parameter $\w^i_k$ is locally updated through a batch of samplings. 
% This batched sampling update continues for $N$ iterations for every given policy $\pi_{\bm{\theta}_t}$.
To evaluate the current policy in the {\em average total reward setting}, the TD-error $\delta_{k,\tau}^{i}$ for objective $i$ at iteration $k$ using sample $\tau$ is as follows:
\begin{align}
&\mu^{i}_{k,\tau} = (1-\beta)\mu^{i}_{k,\tau-1}+\beta r^{i}_{k,\tau} \label{eq:critic_avg1},\\
&\delta^{i}_{k,\tau} = r^{i}_{k,\tau}-\mu^{i}_{k,\tau}+\bm{\phi}^{\top}(s_{k,\tau+1})\w^{i}_k-\bm{\phi}^{\top}(s_{k,\tau})\w^{i}_k. \!\!\! \label{eq:critic_avg2}
\end{align}
On the other hand, to evaluate the current policy under the {\em discounted total reward} setting, the TD-error $\delta_{k,\tau}^{i}$ for objective $i$ at iteration $k$ using sample $\tau$ is as follows:
\begin{equation}
    \delta^{i}_{k,\tau}= r^{i}_{k,\tau}+\gamma^{i}\bm{\phi}^{\top}(s_{k,\tau+1})\w^{i}_k-\bm{\phi}^{\top}(s_{k,\tau})\w^{i}_k.
\label{eq:critic_disc}
\end{equation}
% For each objective $i\in\{1,\cdots,M\}$, the parameter is locally updated by following rules:
% \begin{align}
% \mu^{i}_{k,\tau+1}&=(1-\beta)\mu^{i}_{k,\tau}+\beta r^{i}_{k,\tau+1} \label{eq: mu_i} \\
% \delta^{i}_{k,\tau}&= r^{i}_{k,\tau+1}-\mu^{i}_{k,\tau}+\phi^{T}(s_{k,\tau+1})w^{i}_k-\phi^{T}(s_{k,\tau})w^{i}_k \\
% w^i_k&=w^i_k+\frac{\beta}{M}\sum_{\tau=0}^{M-1}\delta^{i}_{k,\tau}\cdot\phi(s_{k,\tau}), \label{eq: tilde_w_i}
% \end{align}
% where $\beta>0$ is the step-size of the critic step, $\mu^{i}_{k,\tau}$ is the estimate of the long-term return of objective $i$, and $\delta^{i}_{k,\tau}$ is the local TD-error for objective $i$ at iteration $k$ using sample $\tau$.

\begin{algorithm}[t!]
  \SetKwInOut{Input}{Input}
  \SetKwInOut{Output}{Output}
  \SetKwFor{ParFor}{for}{do in parallel}{end for}
  \Input{$s_0$, $\bm{\theta}_0$, $\Phi$, $\eta_t$, actor step size $\alpha$, actor iteration $T$, actor batch size $B$}
  \BlankLine
  \For{$t=1,\cdots,T$}{

	\ul{\textbf{Critic Step:}} $\w_t,s_{t,0}\leftarrow$ Algorithm~\ref{alg: critic}($s_{t-1, B}, \bm{\theta}_t$)
 
  	\For{$l=1,\cdots,B$}{
	Execute action $a_{t,l}\sim\pi_{\bm{\theta}_t}(\cdot|s_{t,l})$
 
	Observe state $s_{t,l+1}$ and reward vector $\bm{r}_{t,l+1}$
 
    \ParFor{$i\in [M]$}{
	$\bullet$ \textit{Setting I: Average Reward:} 

        Update $\mu_{t,l}^i$, $\delta^i_{t,l}$ by Eqs.~(\ref{eq:actor_avg1}),(\ref{eq:actor_avg2}), respectively

        \BlankLine
        
        $\bullet$ \textit{Setting II: Discounted Reward:} 

        Update $\delta^i_{t,l}$ by Eq.~(\ref{eq:actor_disc})
    }
  	}

    \ul{\textbf{Actor Step:}}
    
    \ParFor{$i\in[M]$}{
        $\g_t^i = {1\over B}\sum_{l=1}^{B}\delta_{t,l}^i\cdot\bm{\psi}_{t,l}^i$
    }
    $\hat{\bm{\lambda}}^*_t\leftarrow$ Solver($\g^i_t$) to Problem (\ref{eq:qp})
    
    Update $\bm{\lambda}_t$ by Eq.~(\ref{eq:lambda}). Let $\g_t = \sum_{i=1}^M \lambda_t^i\cdot \g_t^i$.
    
    $\bm{\theta}_{t+1}= \bm{\theta}_t+\alpha\cdot \g_t$
  }
  \Output{$\bm{\theta}_{\hat{T}}$ with $\hat{T}$ chosen uniformly from $\{1,\cdots,T\}$}
  \caption{The overall \policy algorithmic framework.}
  \label{alg: actor}
\end{algorithm}

\textbf{2) The Actor Step:} In the actor step, we first compute individual gradient descent directions $\g_t^i$ via the TD-error of each objective, then compute a common gradient descent direction $\g_t$ to update the policy parameter $\bm{\theta}_t$.
First, according to Lemma~\ref{the: pol_gra}, to obtain individual gradient descent directions, we approximate the advantage function by the TD-error for each objective $i\in[M]$.
Similar to the critic step, for the {\em average total reward} setting, the TD-error for objective $i$ at time $t$ using sample $l$ can be computed as:
\begin{align}
&\mu^{i}_{t,l}=(1-\alpha)\mu^{i}_{t,l}+\alpha r^{i}_{t,l} \label{eq:actor_avg1},\\
&\delta^{i}_{t,l}= r^{i}_{t,l}-\mu^{i}_{t,l}+\bm{\phi}^{\top}(s_{t,l+1})\w^{i}_t-\bm{\phi}^{\top}(s_{t,l})\w^{i}_t.\label{eq:actor_avg2}
\end{align}
For the {\em discounted total reward} setting, the TD-error for objective $i$ at time $t$ using sample $l$ can be computed as:
\begin{equation}
    \delta^{i}_{t,l}= r^{i}_{t,l}+\gamma^{i}\bm{\phi}^{\top}(s_{t,l+1})\w^{i}_t-\bm{\phi}^{\top}(s_{t,l})\w^{i}_t.
\label{eq:actor_disc}
\end{equation}
Next, we compute an estimated weight vector $\hat{\bm{\lambda}}^*_t$ by solving the following quadratic programming problem:
\begin{equation} 
\min_{\bm{\lambda}_t\in\mathbb{R}^M}\,\, \| \sum_{i=1}^M\lambda_t^i\cdot \g_t^i \|_2^2\quad
    \text{s.t. } \bm{\lambda}_t\geq \bm{0}, \,\, |\bm{\lambda}_t|_1=1.
\label{eq:qp}
\end{equation}
Then, we update $\bm{\lambda}_t$ by using a momentum coefficient $\eta_t\in[0,1)$, which is given by
\begin{equation}
    \bm{\lambda}_t = (1-\eta_t)\bm{\lambda}_{t-1}+\eta_t\hat{\bm{\lambda}}^*_t.
\label{eq:lambda}
\end{equation}
The complete \policy algorithm is shown in Algorithm~\ref{alg: actor}.

\section{Pareto-Stationary Convergence and Sample Complexity Analysis for \policy}

In this section, we first analyze the convergence and sample complexity of the critic of \policy in Section~\ref{sec:critic}.
Based on these results, we establish the Pareto stationary convergence and sample complexity of \policy in Section~\ref{sec:moac} for both average total reward and discounted total reward settings.
Due to space limitations, we relegate all proofs to the Appendix.

%%% Temporally put here. Will take a more careful look later.

As presented in Section~\ref{sec:model}, \policy is parameterized with $\bm{\theta}\in\mathbb{R}^{d_1}$.
Recall $\bm{\psi}_{\bm{\theta}}(s,a)=\nabla_{\bm{\theta}}\log \pi_{\bm{\theta}}(a|s)$ for any given state-action pair $(s,a)$. 
We state the following assumptions needed for convergence analysis:
\begin{assumption}
For any two policy parameters $\bm{\theta},\bm{\theta}'\in \mathbb{R}^{d_1}$, and any state-action pair $(s,a)\in\mathcal{S}\times\mathcal{A}$, there exist positive constants $C_{\bm{\psi}},L >0$ such that the following hold:
\vspace{-1em}
\begin{list}{\labelitemi}{\leftmargin=1em \itemindent=-0.0em \itemsep=-.2em}
\item[(a)] $\|\bm{\psi}_{\bm{\theta}}(s,a)\|_2\le C_{\bm{\psi}}$;
\item[(b)] $\|\nabla_{\bm{\theta}}J^{i}(\bm{\theta})-\nabla_{\bm{\theta}}J^{i}(\bm{\theta}')\|_2\le L_J\|\bm{\theta}-\bm{\theta}'\|_2, \forall i\in[M]$.
\end{list}
\vspace{-1em}
\label{ass:Lip_bou}
\end{assumption}
Assumption~\ref{ass:Lip_bou} requires that the score function is uniformly bounded for any policy and the gradient of each objective function is Lipschitz with respect to the policy parameter. 
This assumption has also been adopted in the analysis of the single-agent actor-critic RL algorithm in \citep{QiuYanYe_21}. 
For the discounted reward setting, the gradient Lipschitz property can be guaranteed through \citep[Assumption~2]{XuWanLia_20}. 
For the average reward setting, Assumption \ref{ass:Lip_bou} can also be satisfied by the class of soft-max policy under Assumption \ref{ass:mdp} as in \citep{GuoHuZha_21}.
% Also, as stated in Lemma~\ref{lemma:tv}, for aperiodic and irreducible Markov chains, there exists constants $\kappa>1$ and $\rho\in(0,1)$ such that $\sup_{s\in\mathcal{S}}\| P(s_t\mid s_0=s)-d_{\bm{\theta}} \|_{TV}\leq \kappa\rho^t$.
\begin{lemma} \label{lem:mixing}
For any policy $\pi_{\bm{\theta}}$, consider an MDP with transition kernel $P(\cdot\mid s,a)$ and stationary distribution $d_{\bm{\theta}}$, there exists constants $\kappa>0$ and $\rho\in(0, 1)$ such that
\begin{equation*}
     \sup_{s\in\mathcal{S}}\| P(s_t\mid s_0=s)-d_{\bm{\theta}} \|_{TV}\leq \kappa\rho^t.
\end{equation*}
\label{ass:tv}
\end{lemma}
\vspace{-20pt}
Lemma~\ref{lem:mixing} characterizes the mixing time of the underlying Markov process and the data sampled in \policy follows such Markovian process.
As stated in \citet{levin2017markov} (Theorem~4.9), Lemma~\ref{ass:tv} always holds for aperiodic and irreducible Markov chains following from Assumption~\ref{ass:mdp}.

\subsection{Theoretical Results of the Critic of \policy} \label{sec:critic}
The critic step of \policy estimates multi-dimensional rewards and outputs $M$ value function estimations based on the same sequences of Markovian samplings.
In the average reward setting, given a policy parameter $\bm{\theta}$, define vector $\b^i_{\bm{\theta}}:=\mathbb{E}_{s\sim d_{\bm{\theta}},a\sim\pi_{\bm{\theta}}}\left[\left(r^i(s,a)-J^i(\bm{\theta})\right)\bm{\phi}(s)\right], \forall i\in[M]$.
Then the fixed point of TD-learning for objective $i$ is $\w^{i,*}_{\bm{\theta}}=-\A_{\pi_{\bm{\theta}}}^{-1}\b^i_{\bm{\theta}}$, where $\A_{\pi_{\bm{\theta}}}$ is defined in Assumption~\ref{ass:feature}(d).
Similarly, in the discounted reward setting, define vector $\b'^i_{\bm{\theta}}:=\mathbb{E}_{s\sim d_{\bm{\theta}},a\sim\pi_{\bm{\theta}}}\left[r^i(s,a)\bm{\phi}(s)\right], \forall i\in[M]$, and we have $\w^{i,*}_{\bm{\theta}}=-\A^{-1}_{\pi_{\bm{\theta}}}\b'^i_{\bm{\theta}}$. Let constant $C_{\A}>\|\A_{\pi_{\bm{\theta}}}\|_F$ where $\|\cdot\|_F$ denotes the Frobenius Norm.
We now state the convergence of the critic step of \policy as follows:

% Due to Assumption \ref{ass:mdp}, by \citep[Theorem~4.9]{LevPer_17}, for aperiodic and irreducible Makrov chains, we can guarantee the following lemma holds:
% \begin{lemma}
% For any given policy $\pi_{\bm{\theta}}$, consider the MDP with policy $\pi_{\theta}$ and transition kernel $P(\cdot|s,a)$. Let $d_{\theta}$ be the stationary distribution of the MDP. There exist constants $\kappa>1$ and $\rho\in(0,1)$ such that
% $\sup_{s\in\mathcal{S}}||P(s_t|s_0=s)-d_{\theta}||_{TV}\le\kappa\rho^{t}, \quad\forall t\ge 0. $
% \label{lem: erg}
% \end{lemma}

% For the discounted setting, for any given policy $\pi_{\theta}$, there exists $\lambda'_{\A_{\pi_{\theta}}}>0$ such that $(\w-\w^{i,*}_{\bm{\theta}})^{\top} \A_{\pi_{\bm{\theta}}}(\w-\w^{i,*}_{\bm{\theta}})\le -\lambda'_{\A_{\pi_{\theta}}}\|\w-\w^{i,*}_{\bm{\theta}}\|^{2}_2$, which is a property commonly used in the literature (e.g., \citet{XuWanLia_20,bhandari2018finite}).

\begin{theorem}{\em
Under Assumptions \ref{ass:mdp}-\ref{ass:Lip_bou}, for both average and discounted settings, let the critic step size $\beta\leq \min\lbrace {\lambda_{\A}\over 8C_{\A}^2}, {4\over\lambda_{\A}}\rbrace$.
Then, for any objective $i\in[M]$, the iterates generated by Algorithm~\ref{alg: critic} satisfy the following finite-time convergence error bound:
\begin{equation}
    \mathbb{E}\big[ \| \w^i_N \!-\! \w^{i,*}_{\bm{\theta}} \|^2_2 \big] \!\leq\! C_1\big( 1\!-\!\cfrac{\beta\lambda_{\A}}{8} \big)^N \!\!+\! \cfrac{C_2C_3({2\over\lambda_{\A}}\!+\!2\beta)}{\lambda_{\A}D},\!\!
\label{eq:critic}
\end{equation}
where $C_1=\| \w^i_0 - \w^{i,*}_{\bm{\theta}} \|^2_2$, $C_2=[1+(\kappa-1)\rho]/(1-\rho)$, and $C_3>0$ is a constant depending on $\A_{\pi_{\bm{\theta}}}$, $\b_{\bm{\theta}}^i$, and $\b_{\bm{\theta}}'^i$. Detailed definitions of $C_3$ are provided in Appendix~\ref{sec:critic_proof}.
}
\label{thm:critic}
\end{theorem}

Compared to many existing works \cite{LakSze_18,doan2018distributed,ZhaLiuLiu_21} in RL algorithm finite-time convergence analysis, the samples in our method are correlated (i.e., Markovian noise) instead of i.i.d. noise, which is equivalent to $\rho=0$. 
Despite the fact that Markovian noise introduces extra bias error seen from term $C_2$, our batching approach with size $D>1$ offer two-fold benefits: 
1) Part of the convergence error can be controlled with increasing $D$ (cf. the second term on the RHS in Eq.~(\ref{eq:critic}); 
2) it allows the use of {\em constant} step size, leading to a better sample complexity comparing to non-batch approach \cite{srikant2019finite,QiuYanYe_21} and faster convergence in practice in general.
% \kevin{Not sure whether this statement is true or not. Markovian data should introduce more error in convergence. Also, I think we should talk about the exponential mixing time assumption in this section.}

Theorem~\ref{thm:critic} immediately implies the following sample complexity results for the critic in \policyns:

\begin{corollary}
For both average and discounted settings, let $N\geq{8\over\beta\lambda_{\A}}\log(2C_1/\epsilon)$ and $D\geq C_2C_3\big( {2\over\lambda_{\A}}+2\beta \big)/(\epsilon\lambda_{\A})$.
It then holds that $\mathbb{E}\big[ \| \w^i_N - \w^{i,*}_{\bm{\theta}} \|^2_2 \big]\leq\epsilon, i\in[M]$, which implies a sample complexity of $\mathcal{O}(\epsilon^{-1}\log(\epsilon^{-1}))$.
\label{coro:critic}
\end{corollary}

\subsection{Theoretical Results of the \policy Framework} \label{sec:moac}
Computing a common descent direction out of $M$ policy gradients is essential in the actor step.
Especially, the quadratic programming problem in Eq.~(\ref{eq:qp}) changes over iterations due to individual gradients $\g^i_t$ computed from stochastic data.
Thus, to analyze the convergence of the actor step, it is essential to quantify the cumulative change of $\bm{\lambda}_t$ over iterations.
By introducing a momentum coefficient $\eta_t$, such cumulative change of $\bm{\lambda}_t$ is quantifiable and controllable.
To present the convergence of our method, we first define the approximation error of the critic component: $\zeta_{\text{approx}}:= \max_{i\in[M]}\max_{\bm{\theta}}\mathbb{E}[|V^i(s) - V_{\w^{i,*}}^i(s)|^2]$.
$\zeta_{\text{approx}}$ becomes zero if the true value functions $\{V^{i}(\cdot)\}_{i\in [M]}$ are in the linear function class; % for any $\bm{\theta}$.
otherwise, such approximation error is inevitable.
We note that the use of $\zeta_{\text{approx}}$ is standard in the literature of RL algorithm convergence analysis (e.g., \citet{XuWanLia_20,bhatnagar2009natural,QiuYanYe_21}).
We now state the convergence of \policy to a Pareto-stationarity neighborhood as follows:

% \begin{theorem}
% \label{thm:moac1}{\em
% Suppose Assumptions \ref{ass:mdp}-\ref{ass:Lip_bou} hold.
% For any policy $\pi_{\bm{\theta}}$ in average reward maximization over $M$ objectives, setting actor step size $\alpha={1\over 3L_J}$ and critic step size $\beta\leq \min\lbrace {\lambda_A\over 32},{4\over\lambda_A} \rbrace$, the Pareto stationary convergence rate is as follows:
% \begin{align*}
%     &\mathbb{E}\big[ \|\nabla_{\bm{\theta}}\bm{J}(\bm{\theta}_{\hat{T}})\bm{\lambda}_{\hat{T}}^*\|^2_2 \big]\leq \cfrac{18L_J r_{\max}}{T}\bigg(1 + \sum_{t=1}^T\eta_t \bigg) \nonumber \\
%     &+\cfrac{60}{T}\sum_{t=1}^T\max_{i\in [M]}\mathbb{E}\big[\|\bm{w}^i_t - \bm{w}^{i,*}_t\|_2^2\big] + \mathcal{O}\bigg(\cfrac{1}{B}\bigg) + \mathcal{O}\big(\zeta_{\text{approx}}\big).
% \end{align*}
% Further, set $T\geq 18L_Jr_{\max}/\epsilon\cdot\max\lbrace 1, \sum_{t=1}^T\eta_t \rbrace$, $\mathbb{E}[\|\bm{w}^i_t - \bm{w}^{i,*}_t\|_2^2]\leq {\epsilon/60}, \forall i\in[M]$, and $B=\Theta(\epsilon^{-1})$, then we have
% \begin{equation*}
%     \mathbb{E}\big[ \|\nabla_{\bm{\theta}}\bm{J}(\bm{\theta}_{\hat{T}})\bm{\lambda}_{\hat{T}}^*\|^2_2 \big]\leq \epsilon + \mathcal{O}\big(\zeta_{\text{approx}}\big),
% \end{equation*}
% with total sample complexity of $\mathcal{O}\left(\epsilon^{-2}\log{(\epsilon^{-1})}\right)$.
% }
% \end{theorem}

\begin{theorem}{\em
Under Assumptions \ref{ass:mdp}-\ref{ass:Lip_bou}, set the critic step size $\beta\leq \min\lbrace {\lambda_{\A}\over 8C_{\A}^2}, {4\over\lambda_{\A}}\rbrace$ and the actor step size $\alpha={1\over 3L_J}$, where the choice of $C_{\A}$ depends on the reward setting.
Then, the iterates generated by Algorithm~\ref{alg: actor} satisfy the following finite-time convergence error bound:
\begin{align*}
     &\mathbb{E}\big[ \|\nabla_{\bm{\theta}}\J(\bm{\theta}_{\hat{T}})\bm{\lambda}_{\hat{T}}^*\|^2_2 \big]\leq \cfrac{18L_Jr_{\max}}{C_4T}\bigg(1 + \sum_{t=1}^T2\eta_t \bigg) + \\
     &\cfrac{12}{T}\sum_{t=1}^T\max_{i\in [M]}\mathbb{E}\big[\|\w^i_t \!-\! \w^{i,*}_t\|_2^2\big] \!+\! \cfrac{C_5(1\!-\!\rho\!+\!4\kappa\rho)}{(1\!-\!\rho)B} \!+\! 12\zeta_{\text{approx}}
\end{align*}
where $\hat{T}$ is uniformly sampled among $\{1,\cdots, T\}$ and (i) for average setting $C_4=1$ and $C_5=48(r_{\max}+R_{\w})^2$; and (ii) for discounted setting $C_4=1-\|\bm{\gamma}\|_\infty$ and $C_5=12(r_{\max}+2R_{\w})^2$.
}
\label{thm:moac1}
\end{theorem}

% \begin{theorem}
% \label{thm:moac1}
% Under Assumptions \ref{ass:mdp}-\ref{ass:Lip_bou} and depending on the reward setting, choose the step sizes $\alpha$ and $\beta$ in \policy as follows:
% (i) \ul{Discounted Reward:} Let $\alpha={1\over 3L_J}$ and $\beta\leq \min\lbrace {\lambda_A\over 32},{4\over\lambda_A} \rbrace$;
% (ii) \ul{Average Reward:} Let $\alpha={1\over 3L_J}$ and $\beta\leq \min\lbrace {\lambda_A\over 32},{4\over\lambda_A} \rbrace$.
% Then, the iterates generated by Algorithm~\ref{alg: actor} satisfy the following finite-time convergence error bound:
% \begin{multline*}
%      \mathbb{E}\big[ \|\nabla_{\bm{\theta}}\J(\bm{\theta}_{\hat{T}})\bm{\lambda}_{\hat{T}}^*\|^2_2 \big]\leq \cfrac{C_4}{T}\bigg(1 + \sum_{t=1}^T\eta_t \bigg) \\
%      +\cfrac{60}{T}\sum_{t=1}^T\max_{i\in [M]}\mathbb{E}\big[\|\w^i_t - \w^{i,*}_t\|_2^2\big] + \cfrac{C_5}{B} + \zeta_{\text{approx}},
% \end{multline*}
% where the constants $C_4$ and $C_5$ are defined as follows depending on the reward setting:

% (i) \ul{Discounted Reward:} 
% \begin{align*}
% &C_4=\cfrac{18L_J r_{\max}}{1-\|\bm{\gamma}\|_\infty}, C_5=\cfrac{40(r_{\max} +R_{\w})^2(1-\rho+2\kappa\rho)}{(1-\rho)};
% \end{align*}
% (ii) \ul{Average Reward:} 
% \begin{align*}
% &C_4=18L_J r_{\max}, C_5=\cfrac{90(r_{\max} +R_{\w})^2(1-\rho+4\kappa\rho)}{(1-\rho)}.
% \end{align*}
% \end{theorem}

% For clarity, terms independent of $T$ in Theorem~\ref{thm:moac1} are presented in order and the complete results are shown in Appendix section.
\begin{remark}
Theorem~\ref{thm:moac1} reveals a trade-off between the policy update direction and the eventual convergence rate governed by the parameter $\eta_t \in [0,1]$.
Specifically, by setting $\eta_t=1$, \policy always uses an optimal common descent direction obtained from solving Problem~\eqref{eq:qp} (cf. Eq.~\eqref{eq:lambda}), which may approach a Pareto-stationary point more directly, but eventually induces a linear cumulative change of $\bm{\lambda}_t$ over iterations that is non-vanishing as $T$ gets large (cf. the first term in the bound on RHS).
%, inducing linear cumulative change of $\bm{\lambda}_t$ over iterations, 
%in which case \policy never converges.
On the other hand, if we let $\eta_t$ to be iteration-dependent, e.g., $\eta_t=t^{-1}$ and $ t^{-2}$, then \policy does not precisely follow the optimal common descent direction obtained from solving Problem~\eqref{eq:qp} in each iteration, but guarantees convergence to a neighborhood of Pareto-stationarity at a rate of $\mathcal{O}(T^{-1}\ln T)$ and $\mathcal{O}(T^{-1})$, respectively.
Another advantage of setting $\eta_t=t^{-1}$ or $t^{-2}$ is that the initial gradient weights use the first batch of samples from environment (i.e., $\eta_1=1$ and thus $\bm{\lambda}_1=\hat{\bm{\lambda}}^*_1$ by Eq.~(\ref{eq:lambda})), enhancing the practicality of our approach.
\end{remark}

\begin{remark}
By setting $\eta_t=0$, \policy reduces to an algorithm with pre-specified gradient weights, which is equivalent to the sacralization approach for solving MOO problems with fixed weights.
Although such an algorithm also achieves a convergence rate of order $\mathcal{O}(T^{-1})$, it is hard to guarantee Pareto optimality with pre-specified weights since the algorithm does not explore the Pareto front. 
On the other hand, with $\eta_t>0$, the incorporation of MGDA-type update in \policy facilitates the exploration of the Pareto front, similar to that of the MGDA method to identify the Pareto front of general MOO problem as demonstrated in \cite{zerbinati2011comparison}.
\end{remark}

\begin{corollary}
{\em
%Following the convergence in Theorem~\ref{thm:moac1} and depending on constants $C_4, C_5$ for different reward settings, 
Under the same conditions as in Theorem \ref{thm:moac1}, given any $\epsilon>0$,
by setting $T\geq 18L_Jr_{\max}/(C_4\epsilon)\cdot\max\lbrace 1, \sum_{t=1}^T2\eta_t \rbrace$, $\mathbb{E}[\|\w^i_t - \w^{i,*}_t\|_2^2]\leq {\epsilon/12}, \forall i\in[M]$, and $B\geq C_5(1-\rho+4\kappa\rho)/(\epsilon-\epsilon\rho)$, we have the following:
\begin{equation*}
    \mathbb{E}\big[ \|\nabla_{\bm{\theta}}\J(\bm{\theta}_{\hat{T}})\bm{\lambda}_{\hat{T}}^*\|^2_2 \big]\leq \epsilon + \mathcal{O}\big(\zeta_{\text{approx}}\big),
\end{equation*}
with total sample complexity of $\mathcal{O}\left(\epsilon^{-2}\log{(\epsilon^{-1})}\right)$.
}
\label{corollary:conv_rate}
\end{corollary}

Note that Theorem~\ref{thm:moac1} and Corollary~\ref{corollary:conv_rate} show the convergence rate and sample complexity of \policy are independent of the number of objectives $M$, and the sample complexity of \policy for MORL is the {\em same} as the state-of-the-art sample complexity for single-objective RL~\citep{XuWanLia_20}.
%This can be understood by the fact that the MGDA-type approach in \policy finds a common direction with more influence from the mean of individual directions, instead of their amount.
\section{Experiments}
In this section, we evaluate our \policy algorithmic framework and compare the performance of \policy with other related state-of-the-art methods on sythetic and real-world datasets.
Due to space limitations, we present part of the experiments here and relegate the rest in the Appendix.

\begin{figure}[t]
\centerline{
\begin{subfigure}[b]{0.24\textwidth}
\includegraphics[trim=0 10 0 0, width=\textwidth]{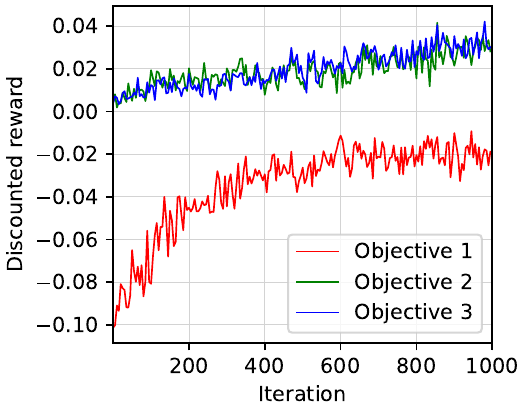}
\caption{}
\end{subfigure}
\hfill
\begin{subfigure}[b]{0.24\textwidth}
\includegraphics[trim=0 10 0 0, width=\textwidth]{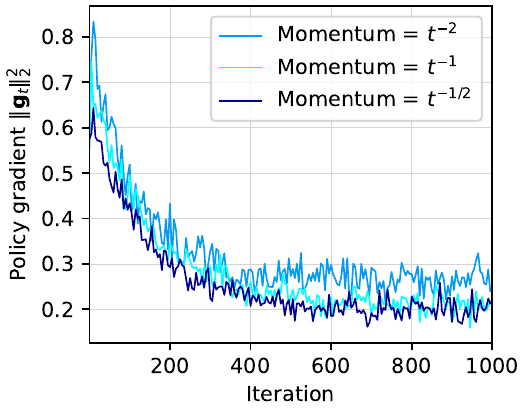}
\caption{}
\end{subfigure}
}
\vspace{-.15in}
\caption{(a) Discounted rewards of three objectives with momentum $\eta_t=t^{-1}$; (b) Squared $\ell_2$-norm of policy gradients with different momentum coefficients.}
\label{fig:momdp}
\vspace{-.25in}
\end{figure}

{\bf 1) Synthetic Data Experiments:}
\textit{1-a) Environment and Setup:} We use an open-source MOMDP environment MO-Gymnasium \citep{Alegre+2022bnaic} to conduct synthetic simulations on environment {\fontfamily{qcr}\selectfont resource-gathering-v0}, which has three reward signals.
We test \policy in the discounted reward setting with momentum coefficient $\eta_t$ chosen from $\lbrace t^{-1/2}, t^{-1}, t^{-2}\rbrace$.
The results are presented in Fig.~\ref{fig:momdp}, where each curve is averaged over $500$ trials.

% \begin{figure}[t!]
% \begin{minipage}{.48\textwidth}
% \includegraphics[trim=0 0 0 0, width=.49\linewidth]{figures/rews_rg.pdf}
% \hfill
% \includegraphics[trim=0 0 0 0, width=.49\linewidth]{figures/grads_rg.pdf}
% \caption{(Left) Discounted rewards of three objectives with momentum $\eta_t=t^{-1}$; (Right) Squared $\ell_2$-norm of policy gradients with different momentum coefficients.}
% \label{fig:momdp}
% \end{minipage}
% \vspace{-.25in}
% \end{figure}

\begin{table*}[t]
\caption{Comparison of our method with baseline methods.}
\centering
\resizebox{.9\textwidth}{!}{
\begin{tabular}{cccccccc}
\toprule
Algorithm & Click$\uparrow$ & Like$\uparrow$(e-2) & Follow$\uparrow$(e-4) & Comment$\uparrow$(e-3) & Forward$\uparrow$(e-3) & Dislike$\downarrow$(e-4) & WatchTime$\uparrow$ \\
\midrule
Behavior-Clone & $0.534$ & $1.231$ & $4.608$ & $3.225^*$ & $1.119^*$ & $2.304$ & $1.285$ \\
\midrule
TSCAC & $0.543^*$  & $1.269$  & $4.535$  & $3.099$  & $1.006$  & $1.342$  & $1.330^*$ \\
      & $1.49\%^*$ & $3.09\%$ &$-1.57\%$ &$-3.92\%$ &$-10.0\%$ &$-41.8\%$ & $3.43\%^*$\\
\midrule
PDPG & $0.539$  & $1.228$  & $4.828^*$  & $3.165$  & $0.919$  & $1.140^*$  & $1.308$ \\
     & $1.02\%$ &$-0.26\%$ & $4.78\%^*$ &$-1.86\%$ &$-17.8\%$ &$-50.5\%^*$ & $1.74\%$\\
\toprule
Ours & $0.539$  & $1.287^*$  & $4.800$  & $3.151$  & $0.897$  & $1.428$  & $1.282$ \\
 (fixed weights)    & $0.98\%$ & $4.54\%^*$ & $4.19\%$ & $-2.29\%$ &$-19.8\%$ &$-38.0\%$ & $-0.27\%$\\
\midrule
Ours & $0.541$  & $\bm{1.312}$  & $\bm{5.070}$  & $\bm{3.266}$  & $1.066$  & $1.486$  & $1.307$ \\
     & $1.30\%$ & $\bm{6.57\%}$ & $\bm{10.0\%}$ & $\bm{1.27\%}$ &$-4.76\%$ &$-35.5\%$ & $1.71\%$\\
\bottomrule
\end{tabular}
}
\label{baselines}
\vspace{-10pt}
\end{table*}

\textit{1-b) Observations:} In Fig.~\ref{fig:momdp}(a), with $\eta_t=t^{-1}$, all objectives are simultaneously improved, and the corresponding policy gradient in Fig.~\ref{fig:momdp}(b) converges. 
%One can understand the policy gradient in our result with the fact that, theoretically, with full-gradients, whenever $\|\g_t\|_2^2<\epsilon$, the current point is $\epsilon$-Pareto stationary and the optimization is interrupted.
Also, as observed in Fig.~\ref{fig:momdp}(b), the policy gradient converges faster with a larger momentum coefficient $\eta_t$, e.g., when $\eta_t=t^{-1/2}$, $\|\g_t\|_2^2$ converges the fastest. 
This is consistent with our theoretical result in Theorem~\ref{thm:moac1} that a larger $\eta_t$-value encourages the policy parameter update direction to follow more closely with the $\hat{\bm{\lambda}}_t^*$-weighted direction, which has a larger descent.
%This indicates that with a larger stride towards optimal descent direction at each iteration, \policy converges faster towards Pareto stationary, which verifies our theoretical results.

{\bf 2) Real-World Data Experiments:}
\textit{2-a) Dataset:} Next, we use a real-world dataset collected from the recommendation logs of the video-sharing mobile app Kuaishou.\footnote{\url{https://kuairand.com/}}
The dataset includes user features and video features, as well as multiple reward signals, such as ``Click,'' ``Like,'' ``Dislike,'' ``WatchTime,'' etc.
The statistic of the dataset is illustrated in Table~\ref{data}.
Specifically, a state corresponds to the event of a video watched by a user, and is represented by the concatenation of user feature and video feature; an action corresponds to a video recommended to a user.
% \begin{table}[h]
% \caption{Data statistic. The reward data is imbalanced, with density of over $98\%$ for the sum of Click and WatchTime.}
% \centering
% \resizebox{.3\textwidth}{!}{
% \begin{tabular}{rlcc}
% \toprule
% \multicolumn{2}{l}{Dimensions}       & Amount & Density \\
% \midrule
% \multicolumn{2}{l}{State}  & 1,218   & - \\
% \midrule
% \multicolumn{2}{l}{Action} & 150    & - \\
% \midrule
% \multirow{7}{*}{\rotatebox[origin=c]{90}{Reward}} & Click     & $254,940$ & $55.25\%$ \\
%        & Like      & $5,190$    & $1.125\%$ \\
%        & Follow    & $203$     & $0.044\%$ \\
%        & Comment   & $1,438$    & $0.312\%$ \\
%        & Forward   & $349$     & $0.076\%$ \\
%        & Dislike   & $213$     & $0.046\%$ \\
%        & WatchTime & $199,122$  & $43.15\%$ \\
% \bottomrule
% \end{tabular}
% }
% \label{data}
% \end{table}
\begin{table}[h]
\vspace{-10pt}
\caption{Data statistic. The reward data is imbalanced, with a density of over $98\%$ for the sum of Click and WatchTime.}
\centering
\resizebox{\linewidth}{!}{
\large
\setlength{\tabcolsep}{3pt}
\begin{tabular}{cccccccc}
\toprule
\multicolumn{8}{l}{State: $1218$ \quad Action: 150} \\
\toprule
&\multicolumn{7}{c}{Reward} \\
\cmidrule(lr){2-8}
& Click & Like & Follow & Comment & Forward & Dislike & WatchTime \\
\midrule
Amount & $254940$ & $5190$ &$203$&$1438$&$349$&$213$&$199122$\\
\midrule
Density &$55.25\%$&$1.125\%$&$0.044\%$&$0.312\%$&$0.076\%$&$0.046\%$&$43.15\%$\\
\bottomrule
\end{tabular}
}
\label{data}
\end{table}
\vspace{-10pt}

\textit{2-b) MORL Environment and Baselines:} 
%To the best of our knowledge, our method is the first online actor-critic method for multi-objective optimization.
To our knowledge, \policy is the first online actor-critic method for MORL.
Thus, there is a lack of direct baseline methods.
To have fair comparisons with other related (offline) MORL algorithms in the literature, we adapt \policy to execute in an off-policy fashion by introducing a behavior policy, which generates actions following from the state-action samples in the dataset.
% \textit{2-c) Baselines:} 
We compare with the following methods.
\vspace{-1em}
\begin{list}{\labelitemi}{\leftmargin=1.5em \itemindent=-0.0em \itemsep=-.2em}
\item \textbf{Behavior-Clone}: A supervised behavior-cloning policy $\pi_{\beta}$ to mimic the recommendation policy in the dataset, which inputs the user state and outputs the video ID.
\item \textbf{PDPG} \citep{CheDuXia_21}: A deterministic policy gradient based actor-critic method that learns Pareto-stationary policy by training networks with model parameters learned from a behavior policy.
This method is the most related to ours, which shares the same goal of finding a Pareto-stationary point for all objectives.
\item \textbf{TSCAC} \citep{CaiXueZha_23}: A two-stage constrained actor-critic approach that optimizes all individual objectives by a set of learned model weights with a focus on optimizing a main objective, while treating other objectives as constraints.
\end{list}
\vspace{-1em}

\textit{2-c) Evaluation:} We adopt Normalised Capped Importance Sampling (NCIS) to evaluate the performances of the methods, which is a standard evaluation approach for off-policy reinforcement learning algorithms \citep{zou2019reinforcement}.
NCIS score quantifies the optimality of a learned policy.
A larger NCIS score implies a better policy for reward maximization.
The definition of NCIS is provided in Section~\ref{metric}.

\textit{2-d) Observations:} The experiment results of baseline comparisons are summarized in Table~\ref{baselines}.
Note that our method and two baselines all start with the same critic and actor parameters initialized for policies that perform worse than Behavior-Clone.
Thus, a negative improvement percentage regarding Behavior-Clone does not imply bad performance on a reward signal.
Based on the result in Table~\ref{baselines}, we have the following observations:
(a) Our method outperforms PDPG in almost all objectives except ``Dislike'' and ``WatchTime''. 
Despite the impact of imbalanced data, our method dominates PDPG in finding a Pareto-efficient policy for multi-objective optimization.
(b) TSCAC outperforms our method in ``Click'', ``Dislike'', and ``WatchTime'', while our method substantially outperforms TSCAC in all other four objectives. 
This is because TSCAC prioritizes WatchTime in optimization, while \policy performs a balanced improvement in all objectives.
(c) Compared to Behavior-Clone, \policy achieves positive improvements in all objectives except ``Forward,'' while other baselines exhibit degraded performance in several objectives. 
TSCAC has negative improvements on ``Follow'', ``Comment'', and ``Forward''; PDPG has negative improvements on ``Like'', ``Comment'', and ``Forward''.

\textit{2-e) Ablation Studies:} 
To show how \policy performs on dynamically deciding a common gradient descent direction, we test another baseline with fixed weights that are initialized by \policy in the first iteration.
The results in Table~\ref{baselines} shows that \policy outperforms the baseline in all the objectives except Dislike.
This comparison indicates a significant performance improvement from the incorporation of momentum-based SMGD.
\section{Conclusion and Future Work}
In this paper, we investigated multi-objective reinforcement learning (MORL) problem by proposing the first MGDA-based actor-critic algorithm called \policyns, which enjoys provable Pareto-stationary convergence and sample complexity guarantees.
Future directions include a generalized model with linear reward signal or nonlinear value function approximation, multi-agent multi-objective reinforcement learning, and decentralized MORL.
% \newpage
\section*{Broader Impact}
Real world applications of our actor-critic framework are broad among various fields.
One typical example is the recommendation system, where our framework provides an architecture with theoretical guarantee.
More applications include automatic driving, robotics, dynamic pricing, etc.
In industry, related methods on MORL have been proposed and applied with different focuses in the last few decades, while our work focuses more on theoretical analysis.
There can be potential societal consequences of our work, but none we feel must be specifically highlighted here.


% \bibliography{MOO/icml2023/refs,MOO/icml2023/refs1}
\bibliography{refs,refs1}

\begin{thebibliography}{50}
\providecommand{\natexlab}[1]{#1}
\providecommand{\url}[1]{\texttt{#1}}
\expandafter\ifx\csname urlstyle\endcsname\relax
  \providecommand{\doi}[1]{doi: #1}\else
  \providecommand{\doi}{doi: \begingroup \urlstyle{rm}\Url}\fi

\bibitem[Abels et~al.(2019)Abels, Roijers, Lenaerts, Now{\'e}, and Steckelmacher]{abels2019dynamic}
Abels, A., Roijers, D., Lenaerts, T., Now{\'e}, A., and Steckelmacher, D.
\newblock Dynamic weights in multi-objective deep reinforcement learning.
\newblock In \emph{International conference on machine learning}, pp.\  11--20. PMLR, 2019.

\bibitem[Alegre et~al.(2022)Alegre, Felten, Talbi, Danoy, Now{\'e}, Bazzan, and da~Silva]{Alegre+2022bnaic}
Alegre, L.~N., Felten, F., Talbi, E.-G., Danoy, G., Now{\'e}, A., Bazzan, A. L.~C., and da~Silva, B.~C.
\newblock {MO-Gym}: A library of multi-objective reinforcement learning environments.
\newblock In \emph{Proceedings of the 34th Benelux Conference on Artificial Intelligence BNAIC/Benelearn 2022}, 2022.

\bibitem[Barrett \& Narayanan(2008)Barrett and Narayanan]{barrett2008learning}
Barrett, L. and Narayanan, S.
\newblock Learning all optimal policies with multiple criteria.
\newblock In \emph{Proceedings of the 25th international conference on Machine learning}, pp.\  41--47, 2008.

\bibitem[Bhatia(2013)]{bhatia2013matrix}
Bhatia, R.
\newblock \emph{Matrix analysis}, volume 169.
\newblock Springer Science \& Business Media, 2013.

\bibitem[Bhatnagar et~al.(2009)Bhatnagar, Sutton, Ghavamzadeh, and Lee]{bhatnagar2009natural}
Bhatnagar, S., Sutton, R.~S., Ghavamzadeh, M., and Lee, M.
\newblock Natural actor--critic algorithms.
\newblock \emph{Automatica}, 45\penalty0 (11):\penalty0 2471--2482, 2009.

\bibitem[Cai et~al.(2023)Cai, Xue, Zhang, Xue, Liu, Zhan, Wang, Zuo, Xie, Zheng, et~al.]{CaiXueZha_23}
Cai, Q., Xue, Z., Zhang, C., Xue, W., Liu, S., Zhan, R., Wang, X., Zuo, T., Xie, W., Zheng, D., et~al.
\newblock Two-stage constrained actor-critic for short video recommendation.
\newblock In \emph{Proceedings of the ACM Web Conference 2023}, pp.\  865--875, 2023.

\bibitem[Chen et~al.(2021{\natexlab{a}})Chen, Du, Xia, and Wang]{CheDuXia_21}
Chen, X., Du, Y., Xia, L., and Wang, J.
\newblock Reinforcement recommendation with user multi-aspect preference.
\newblock In \emph{Proceedings of the Web Conference 2021}, pp.\  425--435, 2021{\natexlab{a}}.

\bibitem[Chen et~al.(2021{\natexlab{b}})Chen, Zhou, Chen, and Zou]{CheZhoChe_21}
Chen, Z., Zhou, Y., Chen, R., and Zou, S.
\newblock Sample and communication-efficient decentralized actor-critic algorithms with finite-time analysis.
\newblock \emph{arXiv preprint arXiv:2109.03699}, 2021{\natexlab{b}}.

\bibitem[Danilova et~al.(2022)Danilova, Dvurechensky, Gasnikov, Gorbunov, Guminov, Kamzolov, and Shibaev]{DanDvuGas_22}
Danilova, M., Dvurechensky, P., Gasnikov, A., Gorbunov, E., Guminov, S., Kamzolov, D., and Shibaev, I.
\newblock Recent theoretical advances in non-convex optimization.
\newblock In \emph{High-Dimensional Optimization and Probability: With a View Towards Data Science}, pp.\  79--163. Springer, 2022.

\bibitem[D{\'e}sid{\'e}ri(2012)]{Des_12}
D{\'e}sid{\'e}ri, J.-A.
\newblock Multiple-gradient descent algorithm (mgda) for multiobjective optimization.
\newblock \emph{Comptes Rendus Mathematique}, 350\penalty0 (5-6):\penalty0 313--318, 2012.

\bibitem[Doan et~al.(2019)Doan, Maguluri, and Romberg]{DoaMagRom_19}
Doan, T., Maguluri, S., and Romberg, J.
\newblock Finite-time analysis of distributed td (0) with linear function approximation on multi-agent reinforcement learning.
\newblock In \emph{International Conference on Machine Learning}, pp.\  1626--1635. PMLR, 2019.

\bibitem[Doan et~al.(2018)Doan, Maguluri, and Romberg]{doan2018distributed}
Doan, T.~T., Maguluri, S.~T., and Romberg, J.
\newblock Distributed stochastic approximation for solving network optimization problems under random quantization.
\newblock \emph{arXiv preprint arXiv:1810.11568}, 2018.

\bibitem[Fernando et~al.(2022)Fernando, Shen, Liu, Chaudhury, Murugesan, and Chen]{FerSheLiu_22}
Fernando, H.~D., Shen, H., Liu, M., Chaudhury, S., Murugesan, K., and Chen, T.
\newblock Mitigating gradient bias in multi-objective learning: A provably convergent approach.
\newblock In \emph{The Eleventh International Conference on Learning Representations}, 2022.

\bibitem[Fliege et~al.(2019)Fliege, Vaz, and Vicente]{FliVazVic_19}
Fliege, J., Vaz, A. I.~F., and Vicente, L.~N.
\newblock Complexity of gradient descent for multiobjective optimization.
\newblock \emph{Optimization Methods and Software}, 34\penalty0 (5):\penalty0 949--959, 2019.

\bibitem[G{\'a}bor et~al.(1998)G{\'a}bor, Kalm{\'a}r, and Szepesv{\'a}ri]{gabor1998multi}
G{\'a}bor, Z., Kalm{\'a}r, Z., and Szepesv{\'a}ri, C.
\newblock Multi-criteria reinforcement learning.
\newblock In \emph{ICML}, volume~98, pp.\  197--205, 1998.

\bibitem[Ge et~al.(2022)Ge, Zhao, Yu, Paul, Hu, Hsieh, and Zhang]{GeZhaYu_22}
Ge, Y., Zhao, X., Yu, L., Paul, S., Hu, D., Hsieh, C.-C., and Zhang, Y.
\newblock Toward pareto efficient fairness-utility trade-off in recommendation through reinforcement learning.
\newblock In \emph{Proceedings of the fifteenth ACM international conference on web search and data mining}, pp.\  316--324, 2022.

\bibitem[Grondman et~al.(2012)Grondman, Busoniu, Lopes, and Babuska]{grondman2012survey}
Grondman, I., Busoniu, L., Lopes, G.~A., and Babuska, R.
\newblock A survey of actor-critic reinforcement learning: Standard and natural policy gradients.
\newblock \emph{IEEE Transactions on Systems, Man, and Cybernetics, Part C (Applications and Reviews)}, 42\penalty0 (6):\penalty0 1291--1307, 2012.

\bibitem[Guo et~al.(2021)Guo, Hu, and Zhang]{GuoHuZha_21}
Guo, X., Hu, A., and Zhang, J.
\newblock Theoretical guarantees of fictitious discount algorithms for episodic reinforcement learning and global convergence of policy gradient methods.
\newblock \emph{arXiv preprint arXiv:2109.06362}, 2021.

\bibitem[Hairi et~al.(2022)Hairi, Liu, and Lu]{HaiLiuLu_22}
Hairi, F., Liu, J., and Lu, S.
\newblock Finite-time convergence and sample complexity of multi-agent actor-critic reinforcement learning with average reward.
\newblock In \emph{International Conference on Learning Representations}, 2022.

\bibitem[Konda \& Tsitsiklis(1999)Konda and Tsitsiklis]{KonTsi_99}
Konda, V. and Tsitsiklis, J.
\newblock Actor-critic algorithms.
\newblock \emph{Advances in neural information processing systems}, 12, 1999.

\bibitem[Kumar et~al.(2019)Kumar, Koppel, and Ribeiro]{kumar2019sample}
Kumar, H., Koppel, A., and Ribeiro, A.
\newblock On the sample complexity of actor-critic for reinforcement learning.
\newblock In \emph{Conference on Neural Information Processing Systems (NeurIPS)}, 2019.

\bibitem[Lakshminarayanan \& Szepesvari(2018)Lakshminarayanan and Szepesvari]{LakSze_18}
Lakshminarayanan, C. and Szepesvari, C.
\newblock Linear stochastic approximation: How far does constant step-size and iterate averaging go?
\newblock In \emph{International Conference on Artificial Intelligence and Statistics}, pp.\  1347--1355. PMLR, 2018.

\bibitem[Levin \& Peres(2017)Levin and Peres]{levin2017markov}
Levin, D.~A. and Peres, Y.
\newblock \emph{Markov chains and mixing times}, volume 107.
\newblock American Mathematical Soc., 2017.

\bibitem[Levine et~al.(2016)Levine, Finn, Darrell, and Abbeel]{levine2016end}
Levine, S., Finn, C., Darrell, T., and Abbeel, P.
\newblock End-to-end training of deep visuomotor policies.
\newblock \emph{The Journal of Machine Learning Research}, 17\penalty0 (1):\penalty0 1334--1373, 2016.

\bibitem[Liu \& Vicente(2021)Liu and Vicente]{liu2021stochastic}
Liu, S. and Vicente, L.~N.
\newblock The stochastic multi-gradient algorithm for multi-objective optimization and its application to supervised machine learning.
\newblock \emph{Annals of Operations Research}, pp.\  1--30, 2021.

\bibitem[Mao et~al.(2016)Mao, Alizadeh, Menache, and Kandula]{mao2016resource}
Mao, H., Alizadeh, M., Menache, I., and Kandula, S.
\newblock Resource management with deep reinforcement learning.
\newblock In \emph{the 15th ACM Workshop on Hot Topics in Networks}, pp.\  50--56, 2016.

\bibitem[Miettinen(1999)]{Mie_99}
Miettinen, K.
\newblock \emph{Nonlinear multiobjective optimization}, volume~12.
\newblock Springer Science \& Business Media, 1999.

\bibitem[Petersen et~al.(2019)Petersen, Yang, Grathwohl, Cockrell, Santiago, An, and Faissol]{petersen2019deep}
Petersen, B.~K., Yang, J., Grathwohl, W.~S., Cockrell, C., Santiago, C., An, G., and Faissol, D.~M.
\newblock Deep reinforcement learning and simulation as a path toward precision medicine.
\newblock \emph{Journal of Computational Biology}, 26\penalty0 (6):\penalty0 597--604, 2019.

\bibitem[Qiu et~al.(2021)Qiu, Yang, Ye, and Wang]{QiuYanYe_21}
Qiu, S., Yang, Z., Ye, J., and Wang, Z.
\newblock On finite-time convergence of actor-critic algorithm.
\newblock \emph{IEEE Journal on Selected Areas in Information Theory}, 2\penalty0 (2):\penalty0 652--664, 2021.

\bibitem[Raghu et~al.(2017{\natexlab{a}})Raghu, Komorowski, Ahmed, Celi, Szolovits, and Ghassemi]{raghu2017deep}
Raghu, A., Komorowski, M., Ahmed, I., Celi, L., Szolovits, P., and Ghassemi, M.
\newblock Deep reinforcement learning for sepsis treatment.
\newblock \emph{arXiv preprint arXiv:1711.09602}, 2017{\natexlab{a}}.

\bibitem[Raghu et~al.(2017{\natexlab{b}})Raghu, Komorowski, Celi, Szolovits, and Ghassemi]{raghu2017continuous}
Raghu, A., Komorowski, M., Celi, L.~A., Szolovits, P., and Ghassemi, M.
\newblock Continuous state-space models for optimal sepsis treatment: a deep reinforcement learning approach.
\newblock In \emph{Machine Learning for Healthcare Conference}, pp.\  147--163. PMLR, 2017{\natexlab{b}}.

\bibitem[Roijers et~al.(2018)Roijers, Steckelmacher, and Now{\'e}]{roijers2018multi}
Roijers, D.~M., Steckelmacher, D., and Now{\'e}, A.
\newblock Multi-objective reinforcement learning for the expected utility of the return.
\newblock In \emph{Proceedings of the Adaptive and Learning Agents workshop at FAIM}, volume 2018, 2018.

\bibitem[Sener \& Koltun(2018)Sener and Koltun]{SenKol_18}
Sener, O. and Koltun, V.
\newblock Multi-task learning as multi-objective optimization.
\newblock \emph{Advances in neural information processing systems}, 31, 2018.

\bibitem[Srikant \& Ying(2019)Srikant and Ying]{srikant2019finite}
Srikant, R. and Ying, L.
\newblock Finite-time error bounds for linear stochastic approximation andtd learning.
\newblock In \emph{Conference on Learning Theory}, pp.\  2803--2830. PMLR, 2019.

\bibitem[Stamenkovic et~al.(2022)Stamenkovic, Karatzoglou, Arapakis, Xin, and Katevas]{StaKarAra_22}
Stamenkovic, D., Karatzoglou, A., Arapakis, I., Xin, X., and Katevas, K.
\newblock Choosing the best of both worlds: Diverse and novel recommendations through multi-objective reinforcement learning.
\newblock In \emph{Proceedings of the Fifteenth ACM International Conference on Web Search and Data Mining}, pp.\  957--965, 2022.

\bibitem[Sutton \& Barto(2018)Sutton and Barto]{SutBar_18}
Sutton, R.~S. and Barto, A.~G.
\newblock \emph{Reinforcement learning: An introduction}.
\newblock MIT press, 2018.

\bibitem[Sutton et~al.(1999)Sutton, McAllester, Singh, Mansour, et~al.]{SutMcASin_99}
Sutton, R.~S., McAllester, D.~A., Singh, S.~P., Mansour, Y., et~al.
\newblock Policy gradient methods for reinforcement learning with function approximation.
\newblock In \emph{NIPs}, volume~99, pp.\  1057--1063. Citeseer, 1999.

\bibitem[Theocharous et~al.(2015)Theocharous, Thomas, and Ghavamzadeh]{theocharous2015personalized}
Theocharous, G., Thomas, P.~S., and Ghavamzadeh, M.
\newblock Personalized ad recommendation systems for life-time value optimization with guarantees.
\newblock In \emph{the 24th International Joint Conference on Artificial Intelligence}, 2015.

\bibitem[Tsitsiklis \& Van~Roy(1999)Tsitsiklis and Van~Roy]{TsiVan_99}
Tsitsiklis, J.~N. and Van~Roy, B.
\newblock Average cost temporal-difference learning.
\newblock \emph{Automatica}, 35\penalty0 (11):\penalty0 1799--1808, 1999.

\bibitem[Van~Moffaert \& Now{\'e}(2014)Van~Moffaert and Now{\'e}]{van2014multi}
Van~Moffaert, K. and Now{\'e}, A.
\newblock Multi-objective reinforcement learning using sets of pareto dominating policies.
\newblock \emph{The Journal of Machine Learning Research}, 15\penalty0 (1):\penalty0 3483--3512, 2014.

\bibitem[Wei et~al.(2022)Wei, Liu, and Ying]{WeiLiuYin_22}
Wei, H., Liu, X., and Ying, L.
\newblock Triple-q: A model-free algorithm for constrained reinforcement learning with sublinear regret and zero constraint violation.
\newblock In Camps-Valls, G., Ruiz, F. J.~R., and Valera, I. (eds.), \emph{Proceedings of The 25th International Conference on Artificial Intelligence and Statistics}, volume 151 of \emph{Proceedings of Machine Learning Research}, pp.\  3274--3307. PMLR, 28--30 Mar 2022.
\newblock URL \url{https://proceedings.mlr.press/v151/wei22a.html}.

\bibitem[Wen et~al.(2023)Wen, Liu, Fedorov, Zhang, Yin, Chu, Hassani, Sun, Liu, Wang, et~al.]{wen2023rankitect}
Wen, W., Liu, K.-H., Fedorov, I., Zhang, X., Yin, H., Chu, W., Hassani, K., Sun, M., Liu, J., Wang, X., et~al.
\newblock Rankitect: Ranking architecture search battling world-class engineers at meta scale.
\newblock \emph{arXiv preprint arXiv:2311.08430}, 2023.

\bibitem[Xu et~al.(2020)Xu, Wang, and Liang]{XuWanLia_20}
Xu, T., Wang, Z., and Liang, Y.
\newblock Improving sample complexity bounds for (natural) actor-critic algorithms.
\newblock \emph{arXiv preprint arXiv:2004.12956}, 2020.

\bibitem[Yang et~al.(2024)Yang, Liu, Liu, Dong, and Momma]{YanLiuLiu_23}
Yang, H., Liu, Z., Liu, J., Dong, C., and Momma, M.
\newblock Federated multi-objective learning.
\newblock 2024.

\bibitem[Yang et~al.(2019)Yang, Sun, and Narasimhan]{yang2019generalized}
Yang, R., Sun, X., and Narasimhan, K.
\newblock A generalized algorithm for multi-objective reinforcement learning and policy adaptation.
\newblock \emph{Advances in neural information processing systems}, 32, 2019.

\bibitem[Zerbinati et~al.(2011)Zerbinati, Desideri, and Duvigneau]{zerbinati2011comparison}
Zerbinati, A., Desideri, J.-A., and Duvigneau, R.
\newblock \emph{Comparison between MGDA and PAES for multi-objective optimization}.
\newblock PhD thesis, INRIA, 2011.

\bibitem[Zhang et~al.(2018)Zhang, Yang, Liu, Zhang, and Basar]{ZhaYanLiu_18}
Zhang, K., Yang, Z., Liu, H., Zhang, T., and Basar, T.
\newblock Fully decentralized multi-agent reinforcement learning with networked agents.
\newblock In \emph{International Conference on Machine Learning}, pp.\  5872--5881. PMLR, 2018.

\bibitem[Zhang et~al.(2021)Zhang, Liu, Liu, Zhu, and Lu]{ZhaLiuLiu_21}
Zhang, X., Liu, Z., Liu, J., Zhu, Z., and Lu, S.
\newblock Taming communication and sample complexities in decentralized policy evaluation for cooperative multi-agent reinforcement learning.
\newblock In \emph{Advances Neural Information Processing Systems (NeurIPS)}, Virtual Event, December 2021.

\bibitem[Zhou et~al.(2022)Zhou, Zhang, Jiang, Zhong, Gu, and Zhu]{ZhoZhaJia_22}
Zhou, S., Zhang, W., Jiang, J., Zhong, W., Gu, J., and Zhu, W.
\newblock On the convergence of stochastic multi-objective gradient manipulation and beyond.
\newblock \emph{Advances in Neural Information Processing Systems}, 35:\penalty0 38103--38115, 2022.

\bibitem[Zou et~al.(2019)Zou, Xia, Ding, Song, Liu, and Yin]{zou2019reinforcement}
Zou, L., Xia, L., Ding, Z., Song, J., Liu, W., and Yin, D.
\newblock Reinforcement learning to optimize long-term user engagement in recommender systems.
\newblock In \emph{Proceedings of the 25th ACM SIGKDD International Conference on Knowledge Discovery \& Data Mining}, pp.\  2810--2818, 2019.

\end{thebibliography}
\bibliographystyle{icml2024}

%%%%%%%%%%%%%%%%%%%%%%%%%%%%%%%%%%%%%%%%%%%%%%%%%%%%%%%%%%%%%%%%%%%%%%%%%%%%%%%
%%%%%%%%%%%%%%%%%%%%%%%%%%%%%%%%%%%%%%%%%%%%%%%%%%%%%%%%%%%%%%%%%%%%%%%%%%%%%%%
% APPENDIX
%%%%%%%%%%%%%%%%%%%%%%%%%%%%%%%%%%%%%%%%%%%%%%%%%%%%%%%%%%%%%%%%%%%%%%%%%%%%%%%
%%%%%%%%%%%%%%%%%%%%%%%%%%%%%%%%%%%%%%%%%%%%%%%%%%%%%%%%%%%%%%%%%%%%%%%%%%%%%%%
\newpage
\appendix
\onecolumn
\section*{Appendix}
In this paper, we use $\|\cdot\|_2, |\cdot|_1, \|\cdot\|_{\infty}$, $\|\cdot\|_F$ to denote $\ell_2$, $\ell_1$, $\ell_{\infty}$, and Frobenius norms respectively, and $\|\cdot\|_{TV}$ for total variance norm. $\langle \cdot,\cdot \rangle$ denotes the inner product. Superscript $i$ in quantity $x$, i.e. $x^{i}$, denotes the $x$ quantity correspond to objective $i\in [M]$. $\lambda(\cdot)$ and $\sigma(\cdot)$ denote the eigenvalues and singular values of the corresponding matrix respectively. All vectors are assumed to be column vector, unless specified. $(\cdot)^{\top}$ is the transpose of an matrix or vector. We use $\mathbf{1}$ to denote all-1 vector with an appropriate dimension.

\section{Experimental Setup and Complementary Results}

\subsection{Synthetic Data}
\textbf{MOMDP Environment.} We conduct synthetic simulations on two environments, described as follows:

\begin{wrapfigure}{R}{0.2\textwidth}
  \includegraphics[trim=0 0 0 0, width=.15\textwidth]{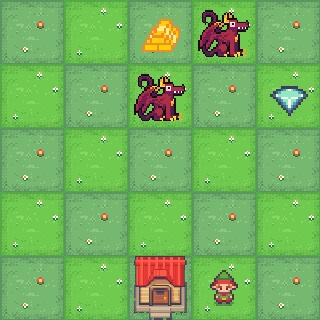}
\caption{Environment: Resource Gathering}
\end{wrapfigure}

Environment {\fontfamily{qcr}\selectfont resource-gathering-v0} \citep{barrett2008learning}:
\vspace{-10pt}
\begin{list}{\labelitemi}{\leftmargin=1.5em \itemindent=0.0em \itemsep=-.2em}
\item \textbf{State space:} $0$ (x coordinate of agent), $1$ (y coordinate of agent), $2$ (flag: gold collected), $3$ (flag: diamond collected)
\item \textbf{Action space:} $0$ (up), $1$ (down), $2$ (left), $3$ (right)
\item \textbf{Reward space:} obj 1: $-1$ (killed by enemy), obj 2: $+1$ (return home with gold), obj 3: $+1$ (return home with diamond)
\item \textbf{Starting state:} The agent starts at the home position with no gold or diamond.
\item \textbf{Episode termination:} When the agent returns home, or when the agent is killed by an enemy.
\end{list}

The FishWood environment is a simple MORL problem in which the agent controls a fisherman which can either fish or go collect wood. In this environment, fishing and collect wood are two conflicting objectives.

\begin{wrapfigure}{R}{0.2\textwidth}
  \includegraphics[trim=0 0 0 0, width=.15\textwidth]{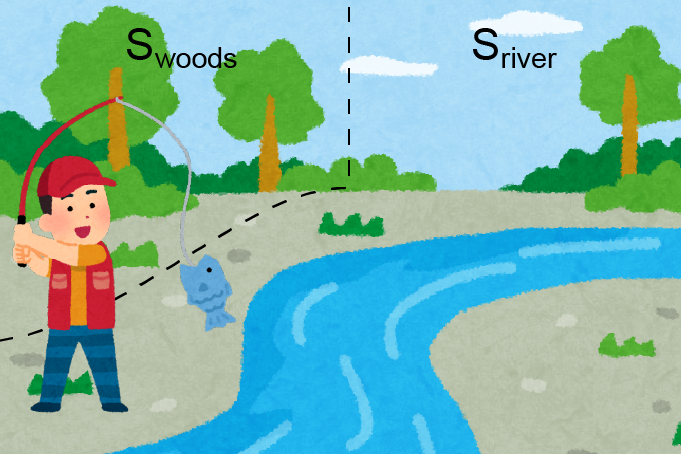}
\caption{Environment: FishWood}
\end{wrapfigure}

Environment {\fontfamily{qcr}\selectfont fishwood-v0} \citep{roijers2018multi}:
\vspace{-10pt}
\begin{list}{\labelitemi}{\leftmargin=1.5em \itemindent=0.0em \itemsep=-.2em}
\item \textbf{State space:} $0$ (fishing), $1$ (in the woods)
\item \textbf{Action space:} $0$ (go fishing), $1$ (go collect wood)
\item \textbf{Reward space:} obj 1: $+1$ (if agent is in the woods, with {\fontfamily{qcr}\selectfont woodproba} probability),\\ obj 2: $+1$ (if the agent is fishing, with {\fontfamily{qcr}\selectfont fishproba} probability)
\item \textbf{Starting state:} Agent starts in the woods
\item \textbf{Episode termination:} The episode ends after {\fontfamily{qcr}\selectfont MAX\_TS}$=200$ steps
\end{list}

\textbf{Simulation results and Observations.}

\begin{figure}[h]
\centering
\begin{minipage}{.7\textwidth}
\includegraphics[trim=0 0 0 0, width=.45\linewidth]{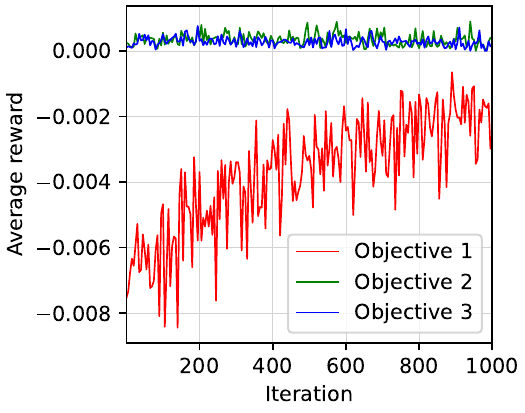}
\hfill
\includegraphics[trim=0 0 0 0, width=.45\linewidth]{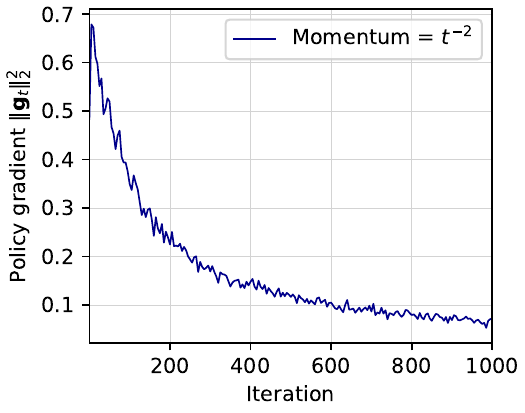}
\caption{Resource Gathering environment. Average rewards of three objectives with momentum $\eta_t=t^{-2}$ (left), and squared $\ell_2$-norm of policy gradients (right).}
\label{fig:momdp3}
\end{minipage}
\end{figure}
Fig.~\ref{fig:momdp3} shows the average rewards of three objectives and corresponding policy gradient in Resource Gathering environment, we can observe that objective 2 and 3 are performing steady and objective 1 is optimized, with a converging policy gradient in the right hand side figure.

\begin{figure}[h]
\centering
\begin{minipage}{.7\textwidth}
\includegraphics[trim=0 0 0 0, width=.45\linewidth]{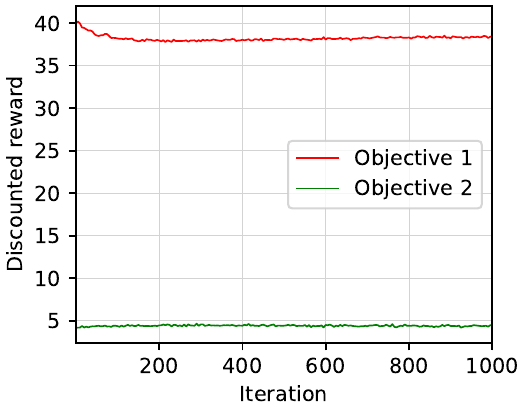}
\hfill
\includegraphics[trim=0 0 0 0, width=.45\linewidth]{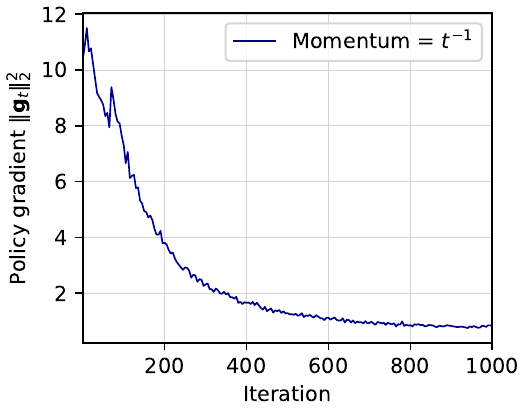}
\caption{FishWood environment. Discounted rewards of two conflicting objectives with momentum $\eta_t=t^{-1}$ (left), and squared $\ell_2$-norm of policy gradients (right).}
\label{fig:momdp2}
\end{minipage}
\end{figure}

Fig.~\ref{fig:momdp2} shows the discounted rewards of two objectives and corresponding policy gradient in FishWood environment, the result shows that two discounted rewards are performing steady with a converging policy gradient.
This is resulted from the fact that two objectives are conflicting, and optimizing any one will sacrifice the other.

\subsection{Real-World Data}
\label{metric}
\textbf{Environment and Setup.} In the dataset, logs provided by the same user are concatenated to form a trajectory in one episode, and a batch of tuple $\lbrace s_t, a_t, \bm{r}_t, s_{t+1}\rbrace$ are sampled at each iteration.
For all the methods, we leverage ADAM to optimize the parameters.
We only experiment on discounted total reward for fair comparison.
For our method, we set the momentum coefficient of gradient weight by $\eta_t=1/t$ (without pre-specifying values, the gradient weights are initialized by the solution to a QP problem regarding the average gradients of the first batch of samples), and set the same gradient weight initialization for all the other methods.

\textbf{Evaluation Metric.} 
Specifically, NCIS score is defined as follows:
\begin{equation*}
    N(\pi) = \cfrac{\sum_{s,a\in D}w(s,a)r(s,a)}{\sum_{s,a\in D}w(s,a)}, \quad w(s,a) = \min\bigg\lbrace C, \cfrac{\pi(a\mid s)}{\pi_{\beta}(a\mid s)} \bigg\rbrace,
\end{equation*}
where $D$ is the dataset, $C$ is a positive constant, and $\pi_{\beta}$ is a behavior policy.

\section{Supporting Lemmas}

\begin{lemma}
Given a policy $\pi_{\bm{\theta}}$, for any objective $i\in[M]$, the TD fixed point for average reward setting $\w_{\bm{\theta}}^{i,*}$ is uniformly bounded, specifically, there exists constant $R_{\w}=4r_{\max}/\lambda_A>0$ such that
\begin{equation*}
    \|\w_{\bm{\theta}}^{i,*}\|\leq R_{\w}, \forall i\in[M].
\end{equation*}
\end{lemma}
\begin{proof}
\begin{align*}
\|\w^{i,*}_{\bm{\theta}}\|_2
&= \|-A_{\pi_{\bm{\theta}}}^{-1}\b^i_{\pi_{\bm{\theta}}}\|_2\\
&= \| -\mathbb{E}_{s\sim d_{\bm{\theta}}(s), s'\sim P(\cdot|s)}[(\bm{\phi}(s')-\bm{\phi}(s))\bm{\phi}^{T}(s)]^{-1}\cdot \mathbb{E}_{s\sim d_{\bm{\theta}},a\sim\pi_{\bm{\theta}}}\left[\bm{\phi}(s)\left(r^i(s,a)-J^i(\bm{\theta})\right)\right] \|_2\\
&\leq \| -\mathbb{E}_{s\sim d_{\bm{\theta}}(s), s'\sim P(\cdot|s)}[(\bm{\phi}(s')-\bm{\phi}(s))\bm{\phi}^{T}(s)]^{-1}\|_2\cdot \|\mathbb{E}_{s\sim d_{\bm{\theta}},a\sim\pi_{\bm{\theta}}}\left[\bm{\phi}(s)\left(r^i(s,a)-J^i(\bm{\theta})\right)\right] \|_2\\
&\overset{\text{(i)}}{=} \cfrac{\|\mathbb{E}_{s\sim d_{\bm{\theta}},a\sim\pi_{\bm{\theta}}}\left[\bm{\phi}(s)\left(r^i(s,a)-J^i(\bm{\theta})\right)\right] \|_2}{{\sigma_{\min}}\left( \| -\mathbb{E}_{s\sim d_{\bm{\theta}}(s), s'\sim P(\cdot|s)}[(\bm{\phi}(s')-\bm{\phi}(s))\bm{\phi}^{T}(s)]\|_2 \right)}\\
&\overset{\text{(ii)}}{\leq} \cfrac{2\|\mathbb{E}_{s\sim d_{\bm{\theta}},a\sim\pi_{\bm{\theta}}}\left[\bm{\phi}(s)\left(r^i(s,a)-J^i(\bm{\theta})\right)\right] \|_2}{\lambda_A\left( -A_{\pi_{\bm{\theta}}} -A_{\pi_{\bm{\theta}}}^\top \right)}\\
&\leq \cfrac{2\cdot\mathbb{E}_{s\sim d_{\bm{\theta}},a\sim\pi_{\bm{\theta}}}\left[\|\bm{\phi}(s)\|_2\cdot \left(|r^i(s,a)|+|J^i(\bm{\theta})|\right)\right] }{\lambda_A}\\
&= \cfrac{4r_{\max}}{\lambda_A},
\end{align*}
where (i) follows from the fact $\|A^{-1}\|=1/\sigma_{\min}(A)$, and (ii) follows from \citet{bhatia2013matrix} (Proposition III 5.1).
\end{proof}

\begin{lemma}
Given a policy $\pi_{\bm{\theta}}$ that maximizes discounted reward, for any objective $i\in[M]$, the optimal value function approximation parameter $\w_{\bm{\theta}}^{i,*}$ is uniformly bounded, specifically, there exists constant $R_{\w}=2r_{\max}/\lambda_A>0$ such that
\begin{equation*}
    \|\w_{\bm{\theta}}^{i,*}\|\leq R_{\w}, \forall i\in[M].
\end{equation*}
\end{lemma}
\begin{proof}
\begin{align*}
\|\w^{i,*}_{\bm{\theta}}\|_2
&= \|-A_{\pi_{\bm{\theta}}}^{-1}\b^i_{\pi_{\bm{\theta}}}\|_2\\
&= \| -\mathbb{E}_{s\sim d_{\bm{\theta}}(s), s'\sim P(\cdot|s)}\left[\left(\gamma\bm{\phi}(s')-\bm{\phi}(s)\right)\bm{\phi}^{T}(s)\right]^{-1}\cdot \mathbb{E}_{s\sim d_{\bm{\theta}},a\sim\pi_{\bm{\theta}}}\left[r^i(s,a)\bm{\phi}(s)\right] \|_2\\
&\leq \| -\mathbb{E}_{s\sim d_{\bm{\theta}}(s), s'\sim P(\cdot|s)}\left[\left(\gamma\bm{\phi}(s')-\bm{\phi}(s)\right)\bm{\phi}^{T}(s)\right]^{-1}\|_2\cdot \|\mathbb{E}_{s\sim d_{\bm{\theta}},a\sim\pi_{\bm{\theta}}}\left[r^i(s,a)\bm{\phi}(s)\right] \|_2\\
&= \cfrac{\|\mathbb{E}_{s\sim d_{\bm{\theta}},a\sim\pi_{\bm{\theta}}}\left[r^i(s,a)\bm{\phi}(s)\right] \|_2}{\| -\mathbb{E}_{s\sim d_{\bm{\theta}}(s), s'\sim P(\cdot|s)}\left[\left(\gamma\bm{\phi}(s')-\bm{\phi}(s)\right)\bm{\phi}^{T}(s)\right]\|_2}\\
&\leq \cfrac{2\|\mathbb{E}_{s\sim d_{\bm{\theta}},a\sim\pi_{\bm{\theta}}}\left[r^i(s,a)\bm{\phi}(s)\right] \|_2}{\lambda_A\left( -A_{\pi_{\bm{\theta}}} -A_{\pi_{\bm{\theta}}}^\top \right)}\\
&\leq \cfrac{2\cdot\mathbb{E}_{s\sim d_{\bm{\theta}},a\sim\pi_{\bm{\theta}}}\left[\|\bm{\phi}(s)\|_2\cdot |r^i(s,a)|\right] }{\lambda_A}\\
&= \cfrac{2r_{\max}}{\lambda_A}.
\end{align*}
\end{proof}

% \begin{lemma}
% For any policy $\pi_{\bm{\theta}}$, consider an MDP with transition kernel $P(\cdot\mid s,a)$ and stationary distribution $d_{\bm{\theta}}$, there exists constants $\kappa>0$ and $\rho\in(0, 1)$ such that
% \begin{equation*}
%      \sup_{s\in\mathcal{S}}\| P(s_t\mid s_0=s)-d_{\bm{\theta}} \|_{TV}\leq \kappa\rho^t.
%  \end{equation*}
%  \label{lemma:tv}
%  \end{lemma}
% As stated in \citet{levin2017markov} (Theorem~4.9), lemma~\ref{lemma:tv} always holds for aperiodic and irreducible Markov chains following from Assumption~\ref{ass:mdp}.
% Similar to Lemma \ref{lemma:tv}, the Markov chain of state-action pair $\lbrace s_t, a_t\rbrace_{t\geq 0}$ with policy $\pi_{\bm{\theta}}$ also has the property of ergodicity, as stated in the following lemma.

\begin{lemma}(\citet{HaiLiuLu_22} Lemma~2)
Let $\nu_{\bm{\theta}}$ denote the stationary distribution of the state-action pairs given policy $\pi_{\bm{\theta}}$, there exists constants $\kappa>0$ and $\rho\in(0, 1)$ such that
\begin{equation*}
    \sup_{s\in\mathcal{S}}\| P(s_t, a_t\mid s_0=s)-\nu_{\bm{\theta}} \|_{TV}\leq \kappa\rho^t.
\end{equation*}
\label{lemma:tv2}
\end{lemma}

\begin{lemma}(\citet{HaiLiuLu_22} Lemma~3)
Suppose Assumption~\ref{ass:feature} holds. Given a policy $\pi_{\bm{\theta}}$, we have the following:
\begin{equation*}
(-\w^{i,*}_{\bm{\theta}})^{\top} \A_{\pi_{\bm{\theta}}}(-\w^{i,*}_{\bm{\theta}})\le -\lambda'_{\A_{\pi_{\theta}}}\|\w^{i,*}_{\bm{\theta}}\|^{2}_2.
\end{equation*}
\end{lemma}

\begin{lemma}(\citet{XuWanLia_20} Theorem~4)
For any $i\in[M]$, consider mini-batched linear stochastic approximation on $\A_{\pi_{\bm{\theta}}}$, $\b_{\bm{\theta}}'^i$ (discounted setting), and $\b_{\bm{\theta}}^i$ (average setting), let $C_{\A}>\|\A_{\pi_{\bm{\theta}}}\|_F$ and $C_{\b}$ denote the upper bound for $\|\b_{\bm{\theta}}^i\|_2$ and $\|\b_{\bm{\theta}}'^i\|_2$, then by setting $\beta\leq \min\lbrace {\lambda_{\A}\over 8C_{\A}^2}, {4\over\lambda_{\A}}\rbrace$ and $D\geq \left({2\over\lambda_{\A}}+2\beta\right){192C_{\A}^2[1+\rho(\kappa-1)]\over(1-\rho)\lambda_{\A}}$ and we have
\begin{equation*}
\mathbb{E}\big[ \| \w^i_N - \w^{i,*}_{\bm{\theta}} \|^2_2 \big] \leq \left( 1 - \cfrac{\beta\lambda_{\A}}{8} \right)^N \cdot\| \w^{i}_0 - \w^{i,*}_{\bm{\theta}} \|^2_2 + \left({2\over\lambda_{\A}}+2\beta\right){192\left(C_{\A}^2R_{\w}^2+C_{\b}^2\right)[1+\rho(\kappa-1)]\over(1-\rho)\lambda_{\A}D}.
\end{equation*}
Further, setting $N\geq{8\over\beta\lambda_{\A}}\log\left(2\| \w^{i}_0 - \w^{i,*}_{\bm{\theta}} \|^2_2/\epsilon\right)$ and $D\geq \left({2\over\lambda_{\A}}+2\beta\right){192\left(C_{\A}^2R_{\w}^2+C_{\b}^2\right)[1+\rho(\kappa-1)]\over\epsilon(1-\rho)\lambda_{\A}}$, we have $\mathbb{E}\big[ \| \w^i_N - \w^{i,*}_{\bm{\theta}} \|^2_2 \big]\leq\epsilon$ with total sample complexity $ND = \mathcal{O}\left(\epsilon^{-1}\log{(\epsilon^{-1})}\right)$.
\label{lemma:xu}
\end{lemma}

\section{Proof of Theorem~\ref{thm:critic}}
\label{sec:critic_proof}
\begin{proof}
The results of Theorem~\ref{thm:critic} follows directly from Lemma~\ref{lemma:xu}, by setting $\A_{\pi_{\bm{\theta}}}:=\mathbb{E}_{s\sim d_{\bm{\theta}}(s), s'\sim P(\cdot|s)}[(\bm{\phi}(s')-\bm{\phi}(s))\bm{\phi}^{\top}(s)]$ and $\b^i_{\bm{\theta}}:=\mathbb{E}_{s\sim d_{\bm{\theta}},a\sim\pi_{\bm{\theta}}}\left[\left(r^i(s,a)-J^i(\bm{\theta})\right)\bm{\phi}(s)\right], \forall i\in[M]$ for the average reward setting, and by setting $\A_{\pi_{\theta}}:=\mathbb{E}_{s\sim d_{\theta}(s), s'\sim P(\cdot|s)}\left[\left(\gamma\bm{\phi}(s')-\bm{\phi}(s)\right)\bm{\phi}^{T}(s)\right]$ and $\b'^i_{\bm{\theta}}:=\mathbb{E}_{s\sim d_{\bm{\theta}},a\sim\pi_{\bm{\theta}}}\left[r^i(s,a)\bm{\phi}(s)\right], \forall i\in[M]$ for the discounted reward setting.

For clarity, we present Theorem~\ref{thm:critic} with some terms simplified as constants, where $C_1=\| \w^i_0 - \w^{i,*}_{\bm{\theta}} \|^2_2$, $C_2=[1+(\kappa-1)\rho]/(1-\rho)$, and $C_3=192\left(C_{\A}^2R_{\w}^2+C_{\b}^2\right)$.
\end{proof}

\section{Proof of Theorem~\ref{thm:moac1}}
% The default norm $\|\cdot\|$ for the proof is $\ell_2$ norm unless otherwise noted. 
For any given $\bm{\theta}$, we denote the gradient matrix to be
\begin{align}
\nabla_{\bm{\theta}}\bm{J}(\bm{\theta})=
\left[\begin{matrix}
\nabla_{\bm{\theta}} J^{1}(\bm{\theta}) &
\nabla_{\bm{\theta}} J^{2}(\bm{\theta}) &
\cdots &
\nabla_{\bm{\theta}} J^{M}(\bm{\theta})
\end{matrix}\right] \in \mathbb{R}^{d_1\times M}.  \nonumber
\end{align}
\begin{proof}
We first present the proof in average reward setting, then we show how to obtain the results in discounted reward setting.
Given $\bm{\theta}\in\mathbb{R}^{d_1}$, $\w\in\mathbb{R}^{d_2}$, $t\geq 0$ and $i\in [M]$, by Lipschitzness in Assumption \ref{ass:Lip_bou},
we have
\begin{align}
J^{i}(\bm{\theta}_{t+1})\ge J^{i}(\bm{\theta}_t)+\left\langle \nabla_{\bm{\theta}}J^{i}(\bm{\theta}_t), \bm{\theta}_{t+1} - \bm{\theta}_t \right\rangle - \cfrac{L_J}{2}\|\bm{\theta}_{t+1} - \bm{\theta}_t\|^2 \label{eq: j_gra_i}
\end{align}
Note that $J^{i}(\bm{\theta})$ is an expected value taken, where the expectation is taken over steady-state distribution induced by policy $\pi_{\bm{\theta}}$. We use $\bm{\lambda}^{*}_t$ to denote the QP solution for using $\{\nabla_{\bm{\theta}}J^{i}(\bm{\theta}_t)\}_{i\in [M]}$, which again is a set of expected vectors. In comparison, $\bm{\lambda}_t$ is the QP solution with momentum for using $\{\g_t^i\}_{i\in [M]}$ as in Algorithm \ref{alg: actor}.

Taking $\bm{\lambda}_t$ weighted summation over Eq. \eqref{eq: j_gra_i}, we have
\begin{align}
    \bm{\lambda}_t^{\top} \bm{J}(\bm{\theta}_{t+1})&\geq \bm{\lambda}_t^\top\bm{J}(\bm{\theta}_t) + \left\langle \nabla_{\bm{\theta}}\bm{J}(\bm{\theta}_t)\bm{\lambda}_t, \bm{\theta}_{t+1} - \bm{\theta}_t \right\rangle - \cfrac{L_J}{2}\|\bm{\theta}_{t+1} - \bm{\theta}_t\|_2^2\nonumber\\
    &= \bm{\lambda}_t^\top\bm{J}(\bm{\theta}_t) + \alpha\left\langle \nabla_{\bm{\theta}}\bm{J}(\bm{\theta}_t)\bm{\lambda}_t, \sum_{j=1}^M \lambda_t^j\cdot\g_t^j \right\rangle - \cfrac{\alpha^2L_J}{2}\|\g_t\|_2^2\nonumber\\
    &= \bm{\lambda}_t^\top\bm{J}(\bm{\theta}_t) + \alpha\left\langle \nabla_{\bm{\theta}}\bm{J}(\bm{\theta}_t)\bm{\lambda}_t, \sum_{j=1}^M \lambda_t^j\cdot\left(\g_t^j - \nabla_{\bm{\theta}}J^j(\bm{\theta}_t) + \nabla_{\bm{\theta}}J^j(\bm{\theta}_t)\right) \right\rangle - \cfrac{\alpha^2L_J}{2}\|\g_t\|_2^2\nonumber\\
    &= \bm{\lambda}_t^\top\bm{J}(\bm{\theta}_t) + \alpha\left\langle \nabla_{\bm{\theta}}\bm{J}(\bm{\theta}_t)\bm{\lambda}_t, \sum_{j=1}^M \lambda_t^j\nabla_{\bm{\theta}}J^j(\bm{\theta}_t) \right\rangle\nonumber\\
    &\hspace{13pt} + \alpha\left\langle \nabla_{\bm{\theta}}\bm{J}(\bm{\theta}_t)\bm{\lambda}_t, \sum_{j=1}^M \lambda_t^j\cdot\left(\g_t^j - \nabla_{\bm{\theta}}J^j(\bm{\theta}_t)\right) \right\rangle- \cfrac{\alpha^2L_J}{2}\|\g_t\|_2^2\nonumber\\
    &= \bm{\lambda}_t^\top\bm{J}(\bm{\theta}_t) + \alpha\left\|\nabla_{\bm{\theta}}\bm{J}(\bm{\theta}_t)\bm{\lambda}_t\right\|^2_2 + \alpha\left\langle \nabla_{\bm{\theta}}\bm{J}(\bm{\theta}_t)\bm{\lambda}_t, \sum_{j=1}^M \lambda_t^j\cdot\left(\g_t^j - \nabla_{\bm{\theta}}J^j(\bm{\theta}_t)\right) \right\rangle - \cfrac{\alpha^2L_J}{2}\|\g_t\|_2^2\nonumber\\
    &\overset{\text{(i)}}{\geq} \bm{\lambda}_t^\top\bm{J}(\bm{\theta}_t) + \cfrac{\alpha}{2}\left\|\nabla_{\bm{\theta}}\bm{J}(\bm{\theta}_t)\bm{\lambda}_t\right\|^2_2 - \cfrac{\alpha}{2}\left\|\sum_{j=1}^M \lambda_t^j\cdot\left(\nabla_{\bm{\theta}}J^j(\bm{\theta}_t)-\g_t^j\right) \right\|^2_2 - \cfrac{\alpha^2L_J}{2}\|\g_t\|_2^2\nonumber\\
    &= \bm{\lambda}_t^\top\bm{J}(\bm{\theta}_t) + \cfrac{\alpha}{2}\left\|\nabla_{\bm{\theta}}\bm{J}(\bm{\theta}_t)\bm{\lambda}_t\right\|^2_2 - \cfrac{\alpha}{2}\left\|\sum_{j=1}^M \lambda_t^j\cdot\left(\nabla_{\bm{\theta}}J^j(\bm{\theta}_t)-\g_t^j\right) \right\|^2_2\nonumber\\
    &\hspace{13pt}- \cfrac{\alpha^2L_J}{2}\left\|\sum_{j=1}^M \lambda_t^j\cdot\left(\g_t^j - \nabla_{\bm{\theta}}J^j(\bm{\theta}_t) + \nabla_{\bm{\theta}}J^j(\bm{\theta}_t)\right)\right\|_2^2\nonumber\\
    &\overset{\text{(ii)}}{\geq} \bm{\lambda}_t^\top\bm{J}(\bm{\theta}_t) + \left(\cfrac{\alpha}{2}-\alpha^2L_J\right)\left\|\nabla_{\bm{\theta}}\bm{J}(\bm{\theta}_t)\bm{\lambda}_t\right\|^2_2 - \left(\cfrac{\alpha}{2}+\alpha^2L_J\right)\left\|\sum_{j=1}^M \lambda_t^j\cdot\left(\nabla_{\bm{\theta}}J^j(\bm{\theta}_t)-\g_t^j\right) \right\|^2_2,
\label{eq:2}
\end{align}

where inequality (i) follows because
% $$\langle \bm{a},\b \rangle\le\frac{1}{2}\|\bm{a}\|^{2}+\frac{1}{2}\|\b\|^{2}.$$
\begin{equation*}
    \left\langle \nabla_{\bm{\theta}}\bm{J}(\bm{\theta}_t)\bm{\lambda}_t, \sum_{j=1}^M \lambda_t^j\cdot\left(\g_t^j-\nabla_{\bm{\theta}}J^j(\bm{\theta}_t)\right) \right\rangle\geq -\cfrac{1}{2}\left\|\nabla_{\bm{\theta}}\bm{J}(\bm{\theta}_t)\bm{\lambda}_t\right\|^2_2 - \cfrac{1}{2}\left\|\sum_{j=1}^M \lambda_t^j\cdot\left(\nabla_{\bm{\theta}}J^j(\bm{\theta}_t)-\g_t^j\right) \right\|^2_2,
\end{equation*}
and inequality (ii) follows because
\begin{equation*}
    \left\|\sum_{j=1}^M \lambda_t^j\cdot\left(\g_t^j - \nabla_{\bm{\theta}}J^j(\bm{\theta}_t) + \nabla_{\bm{\theta}}J^j(\bm{\theta}_t)\right)\right\|_2^2\leq 2\left\|\nabla_{\bm{\theta}}\bm{J}(\bm{\theta}_t)\bm{\lambda}_t\right\|^2_2 + 2\left\|\sum_{j=1}^M \lambda_t^j\cdot\left(\nabla_{\bm{\theta}}J^j(\bm{\theta}_t)-\g_t^j\right) \right\|^2_2.
    % \|\g_t\|_2=\big\| \sum_{i\in[M]}\g^i_t\cdot\lambda_t^i \big\|_2\leq\big\| \sum_{i\in[M]}\g^i_t\big\|_2\leq 4M(r_{\max}+R_{\w}).
\end{equation*}
Taking expectation on both sides of Eq.~(\ref{eq:2}) and conditioning on $\mathcal{F}_t$, we have
\begin{equation*}
    \mathbb{E}\left[ \left\|\nabla_{\bm{\theta}}\bm{J}(\bm{\theta}_t)\bm{\lambda}_t\right\|^2_2 \mid \mathcal{F}_t \right] \leq \cfrac{2\left(\mathbb{E}\left[\bm{\lambda}_t^\top\bm{J}(\bm{\theta}_{t+1})\vert \mathcal{F}_t\right] - \bm{\lambda}_t^\top\bm{J}(\bm{\theta}_t)\right)}{\alpha-2\alpha^2L_J} + \cfrac{\alpha+2\alpha^2L_J}{\alpha-2\alpha^2L_J}\mathbb{E}\left[\left\|\sum_{j=1}^M \lambda_t^j\left(\nabla_{\bm{\theta}}J^j(\bm{\theta}_t)-\g_t^j\right) \right\|^2_2\bigg\vert \mathcal{F}_t\right].
\end{equation*}
By the definitions of $\bm{\lambda}^*_t$ and $\bm{\lambda}_t$, for any time $t$, we have
\begin{equation*}
    \mathbb{E}\left[ \left\|\nabla_{\bm{\theta}}\bm{J}(\bm{\theta}_t)\bm{\lambda}_t^*\right\|^2_2 \mid \mathcal{F}_t \right] \leq \mathbb{E}\left[ \left\|\nabla_{\bm{\theta}}\bm{J}(\bm{\theta}_t)\bm{\lambda}_t\right\|^2_2 \mid \mathcal{F}_t \right].
\end{equation*}
Thus
\begin{equation}
    \mathbb{E}\left[ \left\|\nabla_{\bm{\theta}}\bm{J}(\bm{\theta}_t)\bm{\lambda}_t^*\right\|^2_2 \mid \mathcal{F}_t \right] \leq \cfrac{2\left(\mathbb{E}\left[\bm{\lambda}_t^\top\bm{J}(\bm{\theta}_{t+1})\vert \mathcal{F}_t\right] - \bm{\lambda}_t^\top\bm{J}(\bm{\theta}_t)\right)}{\alpha-2\alpha^2L_J} + \cfrac{\alpha+2\alpha^2L_J}{\alpha-2\alpha^2L_J}\mathbb{E}\left[\left\|\sum_{j=1}^M \lambda_t^j\left(\nabla_{\bm{\theta}}J^j(\bm{\theta}_t)-\g_t^j\right) \right\|^2_2\bigg\vert \mathcal{F}_t\right].
\label{eq: dir_mod}
\end{equation}

\subsection{For the 2nd Term on RHS of Eq.~\eqref{eq: dir_mod}}
Define a notation: $\Delta^j_{\bm{\theta}_t, \w_t^*}=\mathbb{E}_{d_{\bm{\theta}}}\left[\mathbb{E}_{P_{\bm{\theta}}}\left[\delta_{t,l}^j(\w_t^{j,*})\mid(a_{t,l},s_{t,l})\right]\cdot\bm{\psi}^{\bm{\theta}}_{t, l}\right]$.
We first bound the last term on the right hand side of Eq.~\eqref{eq: dir_mod} as follows:
\begin{align}
    &\hspace{13pt}\mathbb{E}\left[\left\|\sum_{j=1}^M \lambda_t^j\left(\nabla_{\bm{\theta}}J^j(\bm{\theta}_t)-\g_t^j\right) \right\|^2_2\bigg\vert \mathcal{F}_t\right]\nonumber\\
    &\leq \mathbb{E}\left[\left(\sum_{j=1}^M \lambda_t^j\left\|\nabla_{\bm{\theta}}J^j(\bm{\theta}_t)-\g_t^j\right\|_2
    \right)^2\bigg\vert \mathcal{F}_t\right]\nonumber\\
    &\leq \mathbb{E}\left[ \left(\sum_{j=1}^M \lambda_t^j \left(\left\| \nabla_{\bm{\theta}}J^j(\bm{\theta}_t) - \Delta^j_{\bm{\theta}_t,\w_t^*}\right\|_2 + \left\| \Delta^j_{\bm{\theta}_t,\w_t^*} -\g^j_{\bm{\theta}_t^*}\right\|_2 + \left\| \g^j_{\bm{\theta}_t^*} - \g_t^j\right\|_2\right) \right)^2 \bigg\vert \mathcal{F}_t \right]\nonumber\\
    &\le 3\mathbb{E}\left[\left(\sum_{j=1}^M \lambda_t^j \left\| \nabla_{\bm{\theta}}J^j(\bm{\theta}_t) - \Delta^j_{\bm{\theta}_t,\w_t^*}\right\|_2  \right)^2\bigg\vert \mathcal{F}_t \right]+ 3\mathbb{E}\left[\left(\sum_{j=1}^M \lambda_t^j \left\| \g^j_{\bm{\theta}_t^*} - \g_t^j\right\|_2 \right)^{2}\bigg\vert  \mathcal{F}_t \right]\nonumber\\
    &\hspace{13pt}+ 3\mathbb{E}\left[ \left(\sum_{j=1}^M \lambda_t^j \cdot\left\| \Delta^j_{\bm{\theta}_t,\w_t^*} -\g^j_{\bm{\theta}_t^*}\right\|_2 \right)^2 \bigg\vert \mathcal{F}_t \right], \label{eq:4} 
    %&\le 2\mathbb{E}\left[ \left(\sum_{j=1}^M \lambda_t^j \left(\left\| \nabla_{\bm{\theta}}J^j(\bm{\theta}_t) - \Delta^j_{\bm{\theta}_t,\w_t^*}\right\|_2 + \left\| \g^j_{\bm{\theta}_t^*} - \g_t^j\right\|_2 \right) \right)^2 \bigg\vert \mathcal{F}_t \right] + 2\mathbb{E}\left[ \left(\sum_{j=1}^M \lambda_t^j \cdot\left\| \Delta^j_{\bm{\theta}_t,\w_t^*} -\g^j_{\bm{\theta}_t^*}\right\|_2 \right)^2 \bigg\vert \mathcal{F}_t \right]\label{eq:4},
\end{align}
where
\begin{align}
    \left\| \nabla_{\bm{\theta}}J^j(\bm{\theta}_t) - \Delta^j_{\bm{\theta}_t,\w_t^*}\right\|^{2}_2&=\left\|\mathbb{E}_{d_{\bm{\theta}}}\left[\mathbb{E}_{P_{\bm{\theta}}}\left[\delta_{t,l}^j\mid(a_{t,l},s_{t,l})\right]\cdot\bm{\psi}^{\bm{\theta}}_{t, l}\right] - \mathbb{E}_{d_{\bm{\theta}}}\left[\mathbb{E}_{P_{\bm{\theta}}}\left[\delta_{t,l}^j(\w_t^{j,*})\mid(a_{t,l},s_{t,l})\right]\cdot\bm{\psi}^{\bm{\theta}}_{t, l}\right]\right\|^{2}_2\nonumber\\
    &= \left\|\mathbb{E}_{d_{\bm{\theta}}}\left[\left(\mathbb{E}_{P_{\bm{\theta}}}\left[\delta_{t,l}^j\mid(a_{t,l},s_{t,l})\right] - \mathbb{E}_{P_{\bm{\theta}}}\left[\delta_{t,l}^j(\w_t^{j,*})\mid(a_{t,l},s_{t,l})\right]\right)\cdot\bm{\psi}^{\bm{\theta}}_{t, l}\right]\right\|^{2}_2\nonumber\\
    &\leq \mathbb{E}_{d_{\bm{\theta}}}\left[\left\|\left(\mathbb{E}_{P_{\bm{\theta}}}\left[\delta_{t,l}^j\mid(a_{t,l},s_{t,l})\right] - \mathbb{E}_{P_{\bm{\theta}}}\left[\delta_{t,l}^j(\w_t^{j,*})\mid(a_{t,l},s_{t,l})\right]\right)\cdot\bm{\psi}^{\bm{\theta}}_{t, l}\right\|^{2}_2\right]\nonumber\\
    &\leq \mathbb{E}_{d_{\bm{\theta}}}\left[\left|\mathbb{E}_{P_{\bm{\theta}}}\left[\delta_{t,l}^j\mid(a_{t,l},s_{t,l})\right] - \mathbb{E}_{P_{\bm{\theta}}}\left[\delta_{t,l}^j(\w_t^{j,*})\mid(a_{t,l},s_{t,l})\right]\right|^{2}\right]\nonumber\\    &=\mathbb{E}_{d_{\bm{\theta}}}\left[\left|\mathbb{E}\left[ V_{\bm{\theta}}^j(s_{t,l+1}) - V_{\bm{\theta}}^j(s_{t,l+1};\w_t^{j,*})\mid(a_{t,l},s_{t,l}) \right] + V_{\bm{\theta}}^j(s_{t,l}) - V_{\bm{\theta}}^j(s_{t,l};\w_t^{j,*})\right|^{2}\right]\nonumber\\
    &\le4 \zeta_{\text{approx}}. \nonumber
\end{align}
We note that $\delta^{j}_{t,l}$ denotes the TD error for objective $j\in [M]$ using the ground truth value functions.
We also remark that the above inequality holds for all $j\in [M]$. As a result, for the first term on the RHS of Eq.~\eqref{eq:4}, we have
\begin{equation}
\mathbb{E}\left[\left(\sum_{j=1}^M \lambda_t^j \left\| \nabla_{\bm{\theta}}J^j(\bm{\theta}_t) - \Delta^j_{\bm{\theta}_t,\w_t^*}\right\|_2  \right)^2\bigg\vert \mathcal{F}_t \right] \nonumber \le \mathbb{E}\left[\left(\sum_{j=1}^M \lambda_t^j 2\sqrt{\zeta_{\text{approx}}} \right)^2\bigg\vert \mathcal{F}_t \right]= 4 \zeta_{\text{approx}} \label{eq:5}
\end{equation}

Furthermore, we have
\begin{align}
    \left\| \g^j_{\bm{\theta}_t^*} - \g_t^j\right\|_2
    &= \left\| \cfrac{1}{B}\sum_{l=0}^{B-1}\left(\delta_{t,l}^j(\w^j_t) - \delta_{t,l}^j(\w^{j,*}_t) \right)\cdot\bm{\psi}^{\bm{\theta}}_{t,l} \right\|_2\nonumber\\
    &=\left\| \cfrac{1}{B}\sum_{l=0}^{B-1}\left( \bm{\phi}(s_{t,l+1}) - \bm{\phi}(s_{t,l}) \right)^{\top}\left( \w^j_t - \w^{j,*}_t \right)\cdot\bm{\psi}^{\bm{\theta}}_{t,l} \right\|_2\nonumber\\
    &\leq \left\| \cfrac{1}{B}\sum_{l=0}^{B-1}\left( \bm{\phi}(s_{t,l+1}) - \bm{\phi}(s_{t,l}) \right)^{\top}\left( \w^j_t - \w^{j,*}_t \right) \right\|_2\nonumber\\
    &\leq \max_{l\in\lbrace 0, \ldots, B-1\rbrace}\left\| \left( \bm{\phi}(s_{t,l+1}) - \bm{\phi}(s_{t,l}) \right)^{\top}\left( \w^j_t - \w^{j,*}_t \right) \right\|_2\nonumber\\
    &\leq 2\cdot\left\| \w^j_t - \w^{j,*}_t \right\|_2. \nonumber
\end{align}
As a result, for the second term on the RHS of Eq.~\eqref{eq:4}, we have
\begin{equation}
\mathbb{E}\left[\left(\sum_{j=1}^M \lambda_t^j \left\| \g^j_{\bm{\theta}_t^*} - \g_t^j\right\|_2 \right)^{2}\bigg\vert  \mathcal{F}_t \right] \le \mathbb{E}\left[\left(\sum_{j=1}^M \lambda_t^j 2\left\| \w^{j}_t-\w^{j,*}_t\right\|_2 \right)^{2}\bigg\vert  \mathcal{F}_t \right] \le 4 \max_{i\in[M]}\mathbb{E} \left[\left\| \w^{i}_t-\w^{i,*}_t\right\|^{2}_2 \bigg\vert  \mathcal{F}_t \right].\label{eq:6}
\end{equation}

Similarly, for the last term in Eq.~(\ref{eq:4}), we have
\begin{equation*}
    \mathbb{E}\bigg[ \bigg(\sum_{j=1}^M \lambda_t^j \cdot\left\| \Delta^j_{\bm{\theta}_t,\w_t^*} -\g^j_{\bm{\theta}_t^*}\right\|_2 \bigg)^2 \bigg\vert \mathcal{F}_t \bigg]
    \le\max_{i\in[M]}\mathbb{E}\bigg[ \bigg(\sum_{j=1}^M \lambda_t^j \cdot\left\| \Delta^i_{\bm{\theta}_t,\w_t^*} -\g^i_{\bm{\theta}_t^*}\right\|_2 \bigg)^2 \bigg\vert \mathcal{F}_t \bigg]
    = \max_{i\in[M]}\mathbb{E}\left[\left\| \Delta^i_{\bm{\theta}_t,\w_t^*} -\g^i_{\bm{\theta}_t^*}\right\|^{2}_2 \bigg\vert \mathcal{F}_t \right].
\end{equation*}
\begin{comment}
\begin{align*}
    &\hspace{13pt}\mathbb{E}\left[ \left(\sum_{j=1}^M \lambda_t^j \cdot\left\| \Delta^j_{\bm{\theta}_t,\w_t^*} -\g^j_{\bm{\theta}_t^*}\right\|_2 \right)^2 \bigg\vert \mathcal{F}_t \right]\\
    &\leq \mathbb{E}\left[ \left(\sum_{j=1}^M \left\| \Delta^j_{\bm{\theta}_t,\w_t^*} -\g^j_{\bm{\theta}_t^*}\right\|_2 \right)^2 \bigg\vert \mathcal{F}_t \right]\\
    &= \mathbb{E}\left[ \sum_{j_1=1}^M\left\| \Delta^{j_1}_{\bm{\theta}_t,\w_t^*} -\g^{j_1}_{\bm{\theta}_t^*} \right\|_2\cdot \sum_{j_2=1}^M\left\| \Delta^{j_2}_{\bm{\theta}_t,\w_t^*} -\g^{j_2}_{\bm{\theta}_t^*} \right\|_2 \bigg\vert\mathcal{F}_t\right]\\
    &= \mathbb{E}\left[ \sum_{j_1=1}^M\sum_{j_2=1}^M \left\| \Delta^{j_1}_{\bm{\theta}_t,\w_t^*} -\g^{j_1}_{\bm{\theta}_t^*} \right\|_2\cdot \left\| \Delta^{j_2}_{\bm{\theta}_t,\w_t^*} -\g^{j_2}_{\bm{\theta}_t^*} \right\|_2 \bigg\vert\mathcal{F}_t\right]\\
    &= \mathbb{E}\left[ \sum_{j=1}^M \left\| \Delta^{j}_{\bm{\theta}_t,\w_t^*} -\g^{j}_{\bm{\theta}_t^*} \right\|_2^2 + \sum_{j_1\neq j_2}\left\| \Delta^{j_1}_{\bm{\theta}_t,\w_t^*} -\g^{j_1}_{\bm{\theta}_t^*} \right\|_2\cdot \left\| \Delta^{j_2}_{\bm{\theta}_t,\w_t^*} -\g^{j_2}_{\bm{\theta}_t^*} \right\|_2 \bigg\vert\mathcal{F}_t\right]\\
    &\leq \mathbb{E}\left[ \sum_{j=1}^M \left\| \Delta^{j}_{\bm{\theta}_t,\w_t^*} -\g^{j}_{\bm{\theta}_t^*} \right\|_2^2 + M\cdot\max_{k\in[M]}\left\| \Delta^{k}_{\bm{\theta}_t,\w_t^*} -\g^{k}_{\bm{\theta}_t^*} \right\|_2^2 \bigg\vert\mathcal{F}_t\right].
\end{align*}
\end{comment}

In addition, for any $j\in [M]$, we have
\begin{align*}
    &\hspace{13pt}\mathbb{E}\left[\left\|\Delta^{j}_{\bm{\theta}_t,\w_t^*} -\g^{j}_{\bm{\theta}_t^*}\right\|_2^2\bigg\vert\mathcal{F}_t\right]\\
    &= \mathbb{E}\left[\left\| \cfrac{1}{B}\sum_{l=0}^{B-1}\delta^j_{t,l}(\w^{j,*}_t)\cdot\bm{\psi}^{\bm{\theta}}_{t,l} - \Delta^j_{\bm{\theta}_t,\w_t^*}\right\|_2^2\bigg\vert\mathcal{F}_t\right]\\
    &= \mathbb{E}\left[ \left\langle \cfrac{1}{B}\sum_{l_1=0}^{B-1}\delta^j_{t,l_1}(\w^{j,*}_t)\cdot\bm{\psi}^{\bm{\theta}}_{t,l_1} - \Delta^j_{\bm{\theta}_t,\w_t^*}, \cfrac{1}{B}\sum_{l_2=0}^{B-1}\delta^j_{t,l_2}(\w^{j,*}_t)\cdot\bm{\psi}^{\bm{\theta}}_{t,l_2} - \Delta^j_{\bm{\theta}_t,\w_t^*} \right\rangle \bigg\vert\mathcal{F}_t\right]\\
    &= \mathbb{E}\left[ \cfrac{1}{B^2}\sum_{l=0}^{B-1}\left\|\delta^j_{t,l}(\w^{j,*}_t)\bm{\psi}^{\bm{\theta}}_{t,l} - \Delta^j_{\bm{\theta}_t,\w_t^*}\right\|_2^2 + \cfrac{1}{B^2}\sum_{l_1\neq l_2}\left\langle\delta^j_{t,l_1}(\w^{j,*}_t)\cdot\bm{\psi}^{\bm{\theta}}_{t,l_1} - \Delta^j_{\bm{\theta}_t,\w_t^*}, \delta^j_{t,l_2}(\w^{j,*}_t)\cdot\bm{\psi}^{\bm{\theta}}_{t,l_2} - \Delta^j_{\bm{\theta}_t,\w_t^*} \right\rangle \bigg\vert\mathcal{F}_t\right]\\
    &\overset{\text{(i)}}{\leq} \cfrac{16}{B}\left(r_{\max} +R_{\w}\right)^2 + \cfrac{1}{B^2}\sum_{l_1\neq l_2}\mathbb{E}\left[ \left\langle\delta^j_{t,l_1}(\w^{j,*}_t)\cdot\bm{\psi}^{\bm{\theta}}_{t,l_1} - \Delta^j_{\bm{\theta}_t,\w_t^*}, \delta^j_{t,l_2}(\w^{j,*}_t)\cdot\bm{\psi}^{\bm{\theta}}_{t,l_2} - \Delta^j_{\bm{\theta}_t,\w_t^*} \right\rangle \bigg\vert\mathcal{F}_t\right]\\
    &= \cfrac{16}{B}(r_{\max} +R_{\w})^2 + \cfrac{2}{B^2}\sum_{l_1<l_2}\mathbb{E}\left[ \left\langle\delta^j_{t,l_1}(\w^{j,*}_t)\cdot\bm{\psi}^{\bm{\theta}}_{t,l_1} - \Delta^j_{\bm{\theta}_t,\w_t^*}, \delta^j_{t,l_2}(\w^{j,*}_t)\cdot\bm{\psi}^{\bm{\theta}}_{t,l_2} - \Delta^j_{\bm{\theta}_t,\w_t^*} \right\rangle \bigg\vert\mathcal{F}_t\right]\\
    &= \cfrac{16}{B}(r_{\max} +R_{\w})^2 + \cfrac{2}{B^2}\sum_{l_1<l_2}\mathbb{E}\left[ \left\langle\delta^j_{t,l_1}(\w^{j,*}_t)\cdot\bm{\psi}^{\bm{\theta}}_{t,l_1} - \Delta^j_{\bm{\theta}_t,\w_t^*}, \mathbb{E}\left[\delta^j_{t,l_2}(\w^{j,*}_t)\cdot\bm{\psi}^{\bm{\theta}}_{t,l_2}\big\vert\mathcal{F}_{t,l_1}\right] - \Delta^j_{\bm{\theta}_t,\w_t^*} \right\rangle \bigg\vert\mathcal{F}_t\right]\\
    &\leq \cfrac{16}{B}(r_{\max} +R_{\w})^2 + \cfrac{2}{B^2}\sum_{l_1<l_2}\mathbb{E}\left[ \left\|\delta^j_{t,l_1}(\w^{j,*}_t)\cdot\bm{\psi}^{\bm{\theta}}_{t,l_1} - \Delta^j_{\bm{\theta}_t,\w_t^*}\right\|_2\cdot\left\|\mathbb{E}\left[\delta^j_{t,l_2}(\w^{j,*}_t)\cdot\bm{\psi}^{\bm{\theta}}_{t,l_2}\big\vert\mathcal{F}_{t,l_1}\right] - \Delta^j_{\bm{\theta}_t,\w_t^*}\right\|_2 \bigg\vert\mathcal{F}_t\right]\\
    &\leq \cfrac{16}{B}(r_{\max} +R_{\w})^2 + \cfrac{2}{B^2}\sum_{l_1<l_2}4\left(r_{\max}+R_{\w}\right)\mathbb{E}\left[\left\|\mathbb{E}\left[\delta^j_{t,l_2}(\w^{j,*}_t)\cdot\bm{\psi}^{\bm{\theta}}_{t,l_2}\big\vert\mathcal{F}_{t,l_1}\right] - \Delta^j_{\bm{\theta}_t,\w_t^*}\right\|_2 \bigg\vert\mathcal{F}_t\right]\\
    &\overset{\text{(ii)}}{\leq} \cfrac{16}{B}(r_{\max} +R_{\w})^2 + \cfrac{2}{B^2}\sum_{l_1<l_2}16(r_{\max} +R_{\w})^2\kappa\rho^{l_2-l_1},
\end{align*}
where (i) follows from the facts that
\begin{align*}
    |\delta^j_{t,l}(\w^{j,*}_t)|
    &= | r^{j}_{t,l+1}-\mu^{j}_{t,l}+\bm{\phi}(s_{t,l+1})^{\top}\w^{j}_t-\bm{\phi}(s_{t,l})^{\top}\w^{j}_t |_1\\
    &\leq | r^{j}_{t,l+1}|+|\mu^{j}_{t,l}|+\|\bm{\phi}(s_{t,l+1})-\bm{\phi}(s_{t,l})\|_2\cdot\|\w^{j}_t\|_2\\
    &\leq 2r_{\max} + 2R_{\w},
\end{align*}
thus, $\|\delta^j_{t,l}(\w^{j,*}_t)\bm{\psi}^{\bm{\theta}}_{t,l}\|_2\leq 2r_{\max} + 2R_{\w}$, and $\Delta^j_{\bm{\theta}_t, \w_t^*}=\mathbb{E}_{d_{\bm{\theta}}}\left[\mathbb{E}_{P_{\bm{\theta}}}\left[\delta_{t,l}^j(\w_t^{j,*})\mid(a_{t,l},s_{t,l})\right]\cdot\bm{\psi}^{\bm{\theta}}_{t, l}\right]\leq 2r_{\max} + 2R_{\w}$,
and (ii) follows from
\begin{align*}
    &\hspace{13pt}\left\|\mathbb{E}\left[\delta^j_{t,l_2}(\w^{j,*}_t)\cdot\bm{\psi}^{\bm{\theta}}_{t,l_2}\big\vert\mathcal{F}_{t,l_1}\right] - \Delta^j_{\bm{\theta}_t,\w_t^*}\right\|_2\\
    &= \left\|\mathbb{E}\left[\delta^j_{t,l_2}(\w^{j,*}_t)\cdot\bm{\psi}^{\bm{\theta}}_{t,l_2}\big\vert\mathcal{F}_{t,l_1}\right] - \mathbb{E}_{d_{\bm{\theta}}}\left[\mathbb{E}_{P_{\bm{\theta}}}\left[\delta_{t,l}^j(\w_t^{j,*})\mid(s_{t,l},a_{t,l})\right]\cdot\bm{\psi}^{\bm{\theta}}_{t, l}\right]\right\|_2\\
    &= \bigg\|\sum_{(s_{t,l_2},a_{t,l_2})}\mathbb{E}_{P_{\bm{\theta}}}\left[\delta_{t,l_2}^j(\w_t^{j,*})\mid(s_{t,l_2},a_{t,l_2})\right]\cdot\bm{\psi}^{\bm{\theta}}_{t, l}\cdot P(s_{t,l_2},a_{t,l_2}\mid\mathcal{F}_{t,l_1})\\
    &\hspace{13pt}-\sum_{(s_{t,l},a_{t,l})}\mathbb{E}_{P_{\bm{\theta}}}\left[\delta_{t,l}^j(\w_t^{j,*})\mid(s_{t,l},a_{t,l})\right]\cdot\bm{\psi}^{\bm{\theta}}_{t, l}\cdot\nu_{\bm{\theta}_t}(s_{t,l},a_{t,l}) \bigg\|_2\\
    &\leq \sum_{(s_{t,l},a_{t,l})}\left\| \mathbb{E}_{P_{\bm{\theta}}}\left[\delta_{t,l}^j(\w_t^{j,*})\mid(s_{t,l},a_{t,l})\right]\cdot\bm{\psi}^{\bm{\theta}}_{t, l} \right\|_2\cdot \left|P^{l_2-l_1}(s_{t,l},a_{t,l}\mid\mathcal{F}_{t,l_1})-\nu_{\bm{\theta}_t}(s_{t,l},a_{t,l})\right|\\
    &\overset{\text{(i)}}{\leq} 4(r_{\max} +R_{\w})\cdot\left\|P^{l_2-l_1}(s,a\mid\mathcal{F}_{t,l_1})-\nu_{\bm{\theta}_t}(s,a) \right\|_{TV}\\
    &\leq 4(r_{\max} +R_{\w})\kappa\rho^{l_2-l_1},
\end{align*}
where (i) follows from Lemma~\ref{lemma:tv2}.

Therefore, for the last term in Eq.~(\ref{eq:4}), we have
\begin{align}
    \mathbb{E}\left[ \left(\sum_{j=1}^M \lambda_t^j \cdot\left\| \Delta^j_{\bm{\theta}_t,\w_t^*} -\g^j_{\bm{\theta}_t^*}\right\|_2 \right)^2 \bigg\vert \mathcal{F}_t \right]
    &\leq \cfrac{16}{B}(r_{\max} +R_{\w})^2 + \cfrac{32}{B^2}\sum_{l_1<l_2}(r_{\max} +R_{\w})^2\kappa\rho^{l_2-l_1}\nonumber\\
    %&= \cfrac{32M}{B}(r_{\max} +R_{\w})^2 + \cfrac{32M}{B^2}(r_{\max} +R_{\w})^2\kappa\sum_{l_1<l_2}\rho^{l_2-l_1}\nonumber\\
    &\leq \cfrac{16}{B}(r_{\max} +R_{\w})^2 + \cfrac{32}{B^2}(r_{\max} +R_{\w})^2\cfrac{2\kappa\rho B}{1-\rho}\nonumber\\
    &= \cfrac{16(r_{\max} +R_{\w})^2(1-\rho+4\kappa\rho)}{(1-\rho)B}.\label{eq:7}
\end{align}
Substituting Eqs.~(\ref{eq:5}), (\ref{eq:6}), (\ref{eq:7}) into Eq.~(\ref{eq:4}) yields the expected gradient bias as follows
\begin{align}
    &\hspace{13pt}\mathbb{E}\left[\left\|\sum_{j=1}^M \lambda_t^j\left(\nabla_{\bm{\theta}}J^j(\bm{\theta}_t)-\g_t^j\right) \right\|^2_2\bigg\vert \mathcal{F}_t\right]\nonumber\\
    &\hspace{13pt} \le 12\zeta_{\text{approx}}+ 12\mathbb{E} \left[\left\| w^{i}_t-w^{i,*}_t\right\|^{2}_2 \bigg\vert  \mathcal{F}_t \right] + \cfrac{48(r_{\max} +R_{\w})^2(1-\rho+4\kappa\rho)}{(1-\rho)B}. \label{eq:8}
    %&\leq \mathbb{E}\left[ \left| \sum_{j=1}^M\lambda^j_t\cdot\left( 2\sqrt{\zeta_{\text{approx}}(t)} + 2\cdot\left\| \w^j_t - \w^{j,*}_t \right\|_2 \right) \right|^2 \bigg\vert\mathcal{F}_t\right] + \cfrac{32M(r_{\max} +R_{\w})^2(1-\rho+2\kappa\rho)}{(1-\rho)B}\nonumber\\
    %&\leq \mathbb{E}\left[ 8M^2{\zeta_{\text{approx}}(t)} + 8M\sum_{j=1}^M\left\|\w^j_t - \w^{j,*}_t\right\|_2^2 \bigg\vert\mathcal{F}_t\right] + \cfrac{32M(r_{\max} +R_{\w})^2(1-\rho+2\kappa\rho)}{(1-\rho)B}\nonumber\\
    %&= 8M^2{\zeta_{\text{approx}}(t)} + 8M\sum_{j=1}^M\mathbb{E}\left[\left\|\w^j_t - \w^{j,*}_t\right\|_2^2 \bigg\vert\mathcal{F}_t\right] + \cfrac{32M(r_{\max} +R_{\w})^2(1-\rho+2\kappa\rho)}{(1-\rho)B}.
\end{align}
Substituting Eq.~(\ref{eq:8}) into Eq.~(\ref{eq: dir_mod}), letting $\alpha=\cfrac{1}{3L_J}$, and taking expectation of $\mathcal{F}_t$ yields
\begin{align}
    \mathbb{E}\left[ \left\|\nabla_{\bm{\theta}}\bm{J}(\bm{\theta}_t)\bm{\lambda}_t^*\right\|^2_2 \right]
    &\leq 18L_J\left(\mathbb{E}\left[\bm{\lambda}_t^\top\bm{J}(\bm{\theta}_{t+1})\right] - \bm{\lambda}_t^\top\bm{J}(\bm{\theta}_t)\right) + 12{\zeta_{\text{approx}}} + 12\max_{j\in [M]}\mathbb{E}\left[\left\|\w^j_t - \w^{j,*}_t\right\|_2^2\right]\nonumber\\
    &\hspace{13pt}+ \cfrac{48(r_{\max} +R_{\w})^2(1-\rho+4\kappa\rho)}{(1-\rho)B}.
\label{eq:9}
\end{align}
% \begin{align}
%     &\hspace{13pt}\mathbb{E}\left[ \left\|\nabla_{\bm{\theta}}\bm{J}(\bm{\theta}_t)\bm{\lambda}_t^*\right\|^2_2 \right]\nonumber\\
%     &\leq \cfrac{2\left(\mathbb{E}\left[\bm{\lambda}_t^\top\bm{J}(\bm{\theta}_{t+1})\vert \mathcal{F}_t\right] - \bm{\lambda}_t^\top\bm{J}(\bm{\theta}_t)\right)}{\alpha-2\alpha^2L_J}\nonumber\\
%     &\hspace{13pt}+ \cfrac{\alpha+2\alpha^2L_J}{\alpha-2\alpha^2L_J}\left[ 8M^2{\zeta_{\text{approx}}(t)} + 8M^2\mathbb{E}\left[\left\|\w^j_t - \w^{j,*}_t\right\|_2^2\right]+ \cfrac{32M(r_{\max} +R_{\w})^2(1-\rho+2\kappa\rho)}{(1-\rho)B} \right]\nonumber\\
%     &= 18L_J\left(\mathbb{E}\left[\bm{\lambda}_t^\top\bm{J}(\bm{\theta}_{t+1})\vert \mathcal{F}_t\right] - \bm{\lambda}_t^\top\bm{J}(\bm{\theta}_t)\right) + 40M^2{\zeta_{\text{approx}}(t)} + 40M^2\mathbb{E}\left[\left\|\w^j_t - \w^{j,*}_t\right\|_2^2\right]\nonumber\\
%     &\hspace{13pt}+ \cfrac{160M(r_{\max} +R_{\w})^2(1-\rho+2\kappa\rho)}{(1-\rho)B}.\label{eq:9}
% \end{align}

\subsection{For the 1st Term on RHS of Eq.~\eqref{eq: dir_mod}}
Let $\hat{T}$ denote a random variable that takes value uniformly random among $\lbrace 1, \ldots, T\rbrace$, then taking average of Eq.~(\ref{eq:9}) over $T$ and we have
\begin{align*}
    \mathbb{E}\left[ \left\|\nabla_{\bm{\theta}}\bm{J}(\bm{\theta}_{\hat{T}})\bm{\lambda}_{\hat{T}}^*\right\|^2_2 \right]
    &= \cfrac{1}{T}\sum_{t=1}^T\mathbb{E}\left[ \left\|\nabla_{\bm{\theta}}\bm{J}(\bm{\theta}_t)\bm{\lambda}_t^*\right\|^2_2 \right]\\
    &\leq \cfrac{18L_J}{T}\sum_{t=1}^T\left(\mathbb{E}\left[\bm{\lambda}_t^\top\bm{J}(\bm{\theta}_{t+1})\right] - \bm{\lambda}_t^\top\bm{J}(\bm{\theta}_t)\right) + \cfrac{12}{T}\sum_{t=1}^T\max_{j\in [M]}\mathbb{E}\left[\left\|\w^j_t - \w^{j,*}_t\right\|_2^2\right]\\
    &\hspace{13pt}+ \cfrac{48(r_{\max} +R_{\w})^2(1-\rho+4\kappa\rho)}{(1-\rho)B} + 12\zeta_{\text{approx}}.
\end{align*}
Specifically,
\begin{align*}
    \sum_{t=1}^T\left(\mathbb{E}\left[\bm{\lambda}_t^\top\bm{J}(\bm{\theta}_{t+1})\right] - \bm{\lambda}_t^\top\bm{J}(\bm{\theta}_t)\right)
    &= \mathbb{E}\left[ \sum_{t=1}^{T-1}(-\bm{\lambda}_{t+1}+\bm{\lambda}_t)^\top\bm{J}(\bm{\theta}_{t+1}) - \bm{\lambda}^\top_1\bm{J}(\bm{\theta}_1) + \bm{\lambda}^\top_T\bm{J}(\bm{\theta}_{T+1}) \right]\\
    &\overset{\text{(i)}}{\leq} \mathbb{E}\left[ \sum_{t=1}^{T-1}|\bm{\lambda}_{t+1}-\bm{\lambda}_t|_{1} \|\bm{J}(\bm{\theta}_{t+1})\|_{\infty} + \|\bm{\lambda}_T\|_{1}\|\bm{J}(\bm{\theta}_{T+1})\|_{\infty} \right]\\
    &\leq r_{\max} + r_{\max}\sum_{t=1}^{T} \mathbb{E}\left[|\bm{\lambda}_{t+1}-\bm{\lambda}_t|_{1}\right]\\
    &= r_{\max} \left(1+ \sum_{t=1}^{T}\eta_t\cdot\mathbb{E}\left[| \hat{\bm{\lambda}}_{t+1}^*-\bm{\lambda}_t|_{1}\right]\right)\\
    &\leq r_{\max} \left(1+ \sum_{t=1}^{T}2\eta_t\right),
    % &= 2r_{\max} + r_{\max}\sum_{t=1}^{T}(1-\eta_t)\| \bm{\lambda}_{t+1}^* - \bm{\lambda}_t^* + \bm{\lambda}_t^* -\bm{\lambda}_t\|_2\\
    % &= 2r_{\max} + r_{\max}\sum_{t=1}^{T}(1-\eta_t)\left[\| \bm{\lambda}_{t+1}^* - \bm{\lambda}_t^*\|_2 + \|\bm{\lambda}_t^* -\bm{\lambda}_t\|_2\right]\\
    % &\leq 2r_{\max} + r_{\max}\sum_{t=1}^{T}(1-\eta_t)\left[4M\alpha (r_{\max}+R_{\w}) + \|\bm{\lambda}_t^* -\bm{\lambda}_t\|_2\right]\\
    % &\leq 2r_{\max} + r_{\max}\sum_{t=1}^T(1-\eta_t).
\end{align*}
where (i) follows from H\"older's Inequality since $1/1+1/\infty=1$. This facilitates the analysis to be $M$-independent in the telescoping process.
Then, we have
\begin{align*}
    \mathbb{E}\left[ \left\|\nabla_{\bm{\theta}}\bm{J}(\bm{\theta}_{\hat{T}})\bm{\lambda}_{\hat{T}}\right\|^2_2 \right]
    &\leq \cfrac{18L_J r_{\max}}{T}\left(1 + \sum_{t=1}^T2\eta_t \right)+\cfrac{12}{T}\sum_{t=1}^T\max_{j\in [M]}\mathbb{E}\left[\left\|\w^j_t - \w^{j,*}_t\right\|_2^2\right]\\
    &\hspace{13pt}+ \cfrac{48(r_{\max} +R_{\w})^2(1-\rho+4\kappa\rho)}{(1-\rho)B} + 12\zeta_{\text{approx}}.
\end{align*}

\subsection{Final Result for Average Reward Setting}
Recalling that $\alpha=\cfrac{1}{3L_J}$ and by letting $T\geq \cfrac{18L_J r_{\max}}{\epsilon}\cdot \max\lbrace 1, \sum_{t=1}^T2\eta_t\rbrace$, $\mathbb{E}\left[\left\|\w^j_t - \w^{j,*}_t\right\|_2^2\right]\leq \cfrac{\epsilon}{12}$ for any objective $j\in[M]$, and $B\geq \cfrac{48(r_{\max} +R_{\w})^2(1-\rho+4\kappa\rho)}{\epsilon}$ yields
\begin{equation*}
    \mathbb{E}\left[ \left\|\bm{\lambda}_{\hat{T}}^\top\nabla_{\bm{\theta}}\bm{J}(\bm{\theta}_{\hat{T}})\right\|^2_2 \right] \leq \epsilon + 60\zeta_{\text{approx}},
\end{equation*}
with a total sample complexity given by
\begin{equation*}
    (B+ND)T=\mathcal{O}\left( \left(\cfrac{1}{\epsilon}+\cfrac{1}{\epsilon}\log\cfrac{1}{\epsilon}\right)\cfrac{1}{\epsilon} \right) = \mathcal{O}\left(\cfrac{1}{\epsilon^2}\log\cfrac{1}{\epsilon}\right).
\end{equation*}

\subsection{Final Result for Discounted Reward Setting}
Similar to the proof in average reward setting, we have
\begin{equation}
    \mathbb{E}\left[ \left\|\nabla_{\bm{\theta}}\bm{J}(\bm{\theta}_t)\bm{\lambda}_t^*\right\|^2_2 \mid \mathcal{F}_t \right] \leq \cfrac{2\left(\mathbb{E}\left[\bm{\lambda}_t^\top\bm{J}(\bm{\theta}_{t+1})\vert \mathcal{F}_t\right] - \bm{\lambda}_t^\top\bm{J}(\bm{\theta}_t)\right)}{\alpha-2\alpha^2L_J} + \cfrac{\alpha+2\alpha^2L_J}{\alpha-2\alpha^2L_J}\mathbb{E}\left[\left\|\sum_{j=1}^M \lambda_t^j\left(\nabla_{\bm{\theta}}J^j(\bm{\theta}_t)-\g_t^j\right) \right\|^2_2\bigg\vert \mathcal{F}_t\right],
\label{eq:10}
\end{equation}
where the last term on the right hand side is bounded by
\begin{align}
&\hspace{13pt}\mathbb{E}\left[\left\|\sum_{j=1}^M \lambda_t^j\left(\nabla_{\bm{\theta}}J^j(\bm{\theta}_t)-\g_t^j\right) \right\|^2_2\bigg\vert \mathcal{F}_t\right] \nonumber\\
 &\le 3\mathbb{E}\left[\left(\sum_{j=1}^M \lambda_t^j \left\| \nabla_{\bm{\theta}}J^j(\bm{\theta}_t) - \Delta^j_{\bm{\theta}_t,\w_t^*}\right\|_2  \right)^2\bigg\vert \mathcal{F}_t \right] \nonumber \\
 &\hspace{13pt}+3\mathbb{E}\left[\left(\sum_{j=1}^M \lambda_t^j \left\| \g^j_{\bm{\theta}_t^*} - \g_t^j\right\|_2 \right)^{2}\bigg\vert  \mathcal{F}_t \right] + 3\mathbb{E}\left[ \left(\sum_{j=1}^M \lambda_t^j \cdot\left\| \Delta^j_{\bm{\theta}_t,\w_t^*} -\g^j_{\bm{\theta}_t^*}\right\|_2 \right)^2 \bigg\vert \mathcal{F}_t \right].
%&\leq \mathbb{E}\left[ \left|\sum_{j=1}^M \lambda_t^j \left(\left\| \nabla_{\bm{\theta}}J^j(\bm{\theta}_t) - \Delta^j_{\bm{\theta}_t,\w_t^*}\right\|_2 + \left\| \g^j_{\bm{\theta}_t^*} - \g_t^j\right\|_2 \right) \right|^2 \bigg\vert \mathcal{F}_t \right]\nonumber\\
%&\hspace{13pt}+ \mathbb{E}\left[ \left|\sum_{j=1}^M \lambda_t^j \cdot\left\| \Delta^j_{\bm{\theta}_t,\w_t^*} -\g^j_{\bm{\theta}_t^*}\right\|_2 \right|^2 \bigg\vert \mathcal{F}_t \right].
\label{eq:11}
\end{align}
Considering the discounted factor $\gamma$, we have
\begin{equation}
    \left\| \nabla_{\bm{\theta}}J^j(\bm{\theta}_t) - \Delta^j_{\bm{\theta}_t,\w_t^*}\right\|_2 \leq 2\sqrt{\zeta_{\text{approx}}},
\label{eq:12}
\end{equation}
and
\begin{equation}
    \left\| \g^j_{\bm{\theta}_t^*} - \g_t^j\right\|_2\leq 2\cdot\left\| \w^j_t - \w^{j,*}_t \right\|_2.
\label{eq:13}
\end{equation}
For the last term in Eq.~(\ref{eq:11}), we have
\begin{equation}
    \mathbb{E}\left[ \left|\sum_{j=1}^M \lambda_t^j \cdot\left\| \Delta^j_{\bm{\theta}_t,\w_t^*} -\g^j_{\bm{\theta}_t^*}\right\|_2 \right|^2 \bigg\vert \mathcal{F}_t \right] \leq \cfrac{4(r_{\max} +2R_{\w})^2(1-\rho+4\kappa\rho)}{(1-\rho)B},
\label{eq:14}
\end{equation}
since the facts
\begin{align*}
    |\delta^j_{t,l}(\w^{j,*}_t)|
    &= | r^{j}_{t,l+1}+\gamma\bm{\phi}(s_{t,l+1})^{\top}\w^{j}_t-\bm{\phi}(s_{t,l})^{\top}\w^{j}_t |_1\\
    &\leq | r^{j}_{t,l+1}|+\|\gamma\bm{\phi}(s_{t,l+1})-\bm{\phi}(s_{t,l})\|_2\cdot\|\w^{j}_t\|_2\\
    &\leq r_{\max} + 2R_{\w},
\end{align*}
thus, $\|\delta^j_{t,l}(\w^{j,*}_t)\bm{\psi}^{\bm{\theta}}_{t,l}\|_2\leq r_{\max} + 2R_{\w}$, and $\Delta^j_{\bm{\theta}_t, \w_t^*}=\mathbb{E}_{d_{\bm{\theta}}}\left[\mathbb{E}_{P_{\bm{\theta}}}\left[\delta_{t,l}^j(\w_t^{j,*})\mid(a_{t,l},s_{t,l})\right]\cdot\bm{\psi}^{\bm{\theta}}_{t, l}\right]\leq r_{\max} + 2R_{\w}$.

Substituting Eqs.~(\ref{eq:12}), (\ref{eq:13}), (\ref{eq:14}) into Eq.~(\ref{eq:11}), we have
\begin{equation}
    \mathbb{E}\left[\left\|\sum_{j=1}^M \lambda_t^j\left(\nabla_{\bm{\theta}}J^j(\bm{\theta}_t)-\g_t^j\right) \right\|^2_2\bigg\vert \mathcal{F}_t\right] \leq 12{\zeta_{\text{approx}}} + 12\max_{j\in[M]}\mathbb{E}\left[\left\|\w^j_t - \w^{j,*}_t\right\|_2^2 \bigg\vert\mathcal{F}_t\right] + \cfrac{12(r_{\max} +2R_{\w})^2(1-\rho+4\kappa\rho)}{(1-\rho)B}.
\label{eq:15}
\end{equation}

Substituting Eq.~(\ref{eq:15}) into Eq.~(\ref{eq:10}), letting $\alpha=\cfrac{1}{3L_J}$, taking expectation of $\mathcal{F}_t$, and taking average of Eq.~(\ref{eq:10}) over $T$ yields
\begin{align*}
    \mathbb{E}\left[ \left\|\nabla_{\bm{\theta}}\bm{J}(\bm{\theta}_{\hat{T}})\bm{\lambda}_{\hat{T}}^*\right\|^2_2 \right]
    &= \cfrac{1}{T}\sum_{t=1}^T\mathbb{E}\left[ \left\|\nabla_{\bm{\theta}}\bm{J}(\bm{\theta}_t)\bm{\lambda}_t^*\right\|^2_2 \right]\\
    &\leq \cfrac{18L_J}{T}\sum_{t=1}^T\left(\mathbb{E}\left[\bm{\lambda}_t^\top\bm{J}(\bm{\theta}_{t+1})\right] - \bm{\lambda}_t^\top\bm{J}(\bm{\theta}_t)\right) + \cfrac{12}{T}\sum_{t=1}^T\max_{j\in[M]}\mathbb{E}\left[\left\|\w^j_t - \w^{j,*}_t\right\|_2^2\right]\\
    &\hspace{13pt}+ \cfrac{4(r_{\max} +2R_{\w})^2(1-\rho+4\kappa\rho)}{(1-\rho)B} + 12\zeta_{\text{approx}},
\end{align*}
where
\begin{align*}
    \sum_{t=1}^T\left(\mathbb{E}\left[\bm{\lambda}_t^\top\bm{J}(\bm{\theta}_{t+1})\right] - \bm{\lambda}_t^\top\bm{J}(\bm{\theta}_t)\right)
    &= \mathbb{E}\left[ \sum_{t=1}^{T-1}(-\bm{\lambda}_{t+1}+\bm{\lambda}_t)^{\top}\bm{J}(\bm{\theta}_{t+1}) - \bm{\lambda}^{\top}_1\bm{J}(\bm{\theta}_1) + \bm{\lambda}^{\top}_T\bm{J}(\bm{\theta}_{T+1}) \right]\\
    &\leq \mathbb{E}\left[ \sum_{t=1}^{T-1}|\bm{\lambda}_{t+1}-\bm{\lambda}_t|_1 \|\bm{J}(\bm{\theta}_{t+1})\|_{\infty} + |\bm{\lambda}_T|_1\|\bm{J}(\bm{\theta}_{T+1})\|_{\infty} \right]\\
    &\leq \sum_{t=1}^{T-1} \left[\eta_t \mathbb{E}\left[|\bm{\lambda}_{t}-\hat{\bm{\lambda}}_t|_1\right]\frac{r_{\max}}{1-\|\bm{\gamma}\|_\infty}\right] + \cfrac{r_{\max}}{1-\|\bm{\gamma}\|_\infty}\\
    &\leq \cfrac{r_{\max}}{1-\|\bm{\gamma}\|_\infty}(1+\sum_{t=1}^{T}2\eta_t).
\end{align*}
Then, we have
\begin{align*}
    \mathbb{E}\left[ \left\|\nabla_{\bm{\theta}}\bm{J}(\bm{\theta}_{\hat{T}})\bm{\lambda}_{\hat{T}}\right\|^2_2 \right]
    &\leq \cfrac{18L_J r_{\max}}{T(1-\|\bm{\gamma}\|_\infty)}(1+2\sum_{t=1}^{T}\eta_t)+ \cfrac{12}{T}\sum_{t=1}^T\max_{j\in[M]}\mathbb{E}\left[\left\|\w^j_t - \w^{j,*}_t\right\|_2^2\right]\\
    &\hspace{13pt}+ \cfrac{12(r_{\max} +2R_{\w})^2(1-\rho+4\kappa\rho)}{(1-\rho)B} + 12\zeta_{\text{approx}}.
\end{align*}
By letting $T\geq \cfrac{18L_J r_{\max}}{\epsilon(1-\|\bm{\gamma}\|_\infty)}\cdot \max\lbrace 1, \sum_{t=1}^T2\eta_t\rbrace$, $\mathbb{E}\left[\left\|\w^j_t - \w^{j,*}_t\right\|_2^2\right]\leq \cfrac{\epsilon}{12}$ for any objective $j\in[M]$, and $B\geq \cfrac{12(r_{\max} +2R_{\w})^2(1-\rho+4\kappa\rho)}{\epsilon}$ yields
\begin{equation*}
    \mathbb{E}\left[ \left\|\bm{\lambda}_{\hat{T}}^\top\nabla_{\bm{\theta}}\bm{J}(\bm{\theta}_{\hat{T}})\right\|^2_2 \right] \leq \epsilon + 12\zeta_{\text{approx}},
\end{equation*}
with total sample complexity given by
\begin{equation*}
    (B+ND)T=\mathcal{O}\left( \left(\cfrac{1}{\epsilon}+\cfrac{1}{\epsilon}\log\cfrac{1}{\epsilon}\right)\cfrac{1}{\epsilon} \right) = \mathcal{O}\left(\cfrac{1}{\epsilon^2}\log\cfrac{1}{\epsilon}\right).
\end{equation*}
\end{proof}

\end{document}